\newif\ifStandalone
\Standalonetrue

\documentclass[square]{ws-rv9x6}
\usepackage{subfigure}   
\usepackage{ws-rv-thm}   
\usepackage{ws-rv-van}   
\makeindex

\DeclareMathOperator{\argmax}{arg\,max}

\ifStandalone
\title{Dealing with Nuisance Parameters using Machine Learning in High Energy Physics: a Review}
\else
\title{Machine Learning in High Energy Physics}
\fi

\ifStandalone
\else
\renewcommand\thechapter{7.\arabic{chapter}}
\fi

\begin{document}

\ifStandalone
\chapter[Dealing with Nuisance Parameters]{Dealing with Nuisance Parameters using Machine Learning in High Energy Physics: a Review}\label{ra_ch1}
\fi
\author[T. Dorigo and P. de Castro Manzano]{ T. Dorigo and P. de Castro Manzano}

\address{Istituto Nazionale di Fisica Nucleare - Sezione di Padova,\\
Via Marzolo 8, 35131 Padova - Italy, \\
tommaso.dorigo@cern.ch\footnote{Corresponding author.} \; pablo.de.castro@cern.ch}

\begin{abstract}
\ifStandalone
In this work we discuss
\fi
the impact of nuisance parameters on the
effectiveness of machine
\ifStandalone
learning in high-energy physics problems, and provide a review of
\fi
techniques that allow to include their effect and reduce their impact in the search for optimal selection criteria and variable transformations. The introduction of nuisance parameters complicates the supervised learning task and its correspondence
with the data analysis goal, due to their contribution degrading the model performances in real data, and the necessary addition of uncertainties in the
resulting statistical inference.
The approaches discussed include nuisance-parameterized
models, modified or adversary losses,
semi-supervised learning approaches, and inference-aware
techniques.
\end{abstract}
\body


\section{Introduction}

\ifStandalone
	Particle
\fi
physics offers a variety of use cases for machine learning techniques. Of these, probably the most common is the use of supervised classification to construct low-dimensional event summaries, which allow to perform statistical inference on the parameters of interest. The learned \textit{summary statistics} --functions of the data that are informative about the relevant properties of the data-- can efficiently combine high-dimensional information from each event into one or a few variables which may be used as the basis of statistical inference. The informational source for this compression are simulated observations produced by a complex generative model; the latter reproduces the chain of physical processes occurring in subnuclear collisions and the subsequent interaction of the produced final state particles with the detection elements.

The fidelity of the event description provided by the simulation is usually limited, due to a variety of factors. These may include imperfections in the modelling of the physical processes employed by the simulation (such as the use of a leading order approximation for the hard scattering process), limited precision in the description of detector response (\textit{e.g.} due to imperfect knowledge of the relevant calibration constants), uncertainty in fundamental physics parameters liable to condition the observations (\textit{e.g.} the mass of a decaying particle), or simply a consequence of the finiteness of the number of simulated observations in a given region of feature space.  To account for these ``known unknowns'', commonly referred to as \textit{nuisance parameters} in statistics literature, the model needs to be enlarged by the inclusion of corresponding variables which are not of direct relevance, yet have to be considered during inference in order to make calibrated statements about the parameters of interest.

Because simulated observations are also at the basis of the construction of the likelihood function or any other estimator employed for the extraction of the wanted information, either directly or through intermediate summary statistics constructed with them, nuisance parameters must be directly included in the statistical model. The inclusion of the effect of nuisances results in the enlargement of confidence intervals on the parameters of interest; nuisance parameters are correspondingly referred to as systematic uncertainties. The effect is not exclusive to parameter estimation problems: nuisance parameters must also be accounted for in hypothesis testing problems, such as when a test of the presence of a new physics signal is performed on data otherwise conforming to the Standard Model hypothesis. The presence of nuisance parameters then causes a reduction of the statistical power of the test. Nuisance parameters are therefore one of main factors limiting the precision and discovery reach of HEP analyses. However, is worth stressing
that nuisance parameters are not intrinsically problematic or negative, but a useful tool
that allows us to model the uncertainties associated to not having
an exact model for the data. 

The mentioned hindrances apply in the same manner to analyses which employ machine learning algorithms to reduce the dimensionality of the data: the effect of nuisance parameters must be accounted for in statistical inference based on summary statistics constructed from the output of the summarising function.
Neither the training loss nor the standard measures of performance for the learning task itself ({\em e.g.} classification or regression) are aligned with the inference goal when the simulated observations depend on additional modelling parameters that are unknown.
\ifStandalone
This review focuses
\fi
on issues arising from the application of machine learning techniques to problems where nuisance parameters are relevant, and the different approaches that have been proposed to overcome the resulting limitations.

\subsection{Probabilistic Classification as Density Ratio Estimation}
\label{sec:classification-as-density-ratio-estimation}

Before delving into the subject matter, it is important to review the relation
between the learning tasks performed in HEP data analysis and the statistical properties of training data. As introduced earlier, machine-learning based
data transformations in HEP are often based on probabilistic classification
models trained with samples from computer simulations of the different
processes. The simplest way to understand a probabilistic classifier is in terms
of probability density ratios.

By training a probabilistic classification model to distinguish samples labelled
by $y=1$ (hereinafter signal simulated samples) and $y=0$ respectively
(background simulated samples), {\em e.g.} a neural network optimising Binary
Cross-Entropy (BCE), we are approximating the density ratio $ r(\boldsymbol{x}) =p(\boldsymbol{x} | y=1)/ p(\boldsymbol{x} | y=0)$ between the signal and background generating distributions
$p(\boldsymbol{x} | y=1)$, $p(\boldsymbol{x} | y=0)$. For example, when using BCE as a loss function, the density ratio $r(\boldsymbol{x})$ can
be approximated using the classification output $c(\boldsymbol{x})$ by virtue of
the following relation:
\begin{equation}
    \frac{c(\boldsymbol{x})}{1-c(\boldsymbol{x})} \approx
    \frac{p(y = 1| \boldsymbol{x})}{p(y = 0| \boldsymbol{x})}
    =  \frac{p(\boldsymbol{x} | y=1)}{p(\boldsymbol{x} | y=0)} \frac{p(y = 1)}{p(y = 0)}
    = r(\boldsymbol{x}) \frac{p(y = 1)}{p(y = 0)}
\label{eq:density_ratio_clf}
\end{equation}
where $p(y = 1)/p(y = 0)$ is independent of $\boldsymbol{x}$, and may be simply
estimated as the ratio between the total number of observations from each
category in the training dataset. The previous approximation becomes an exact
equality only for the best possible classifier, the so-called {\it Bayes optimal
classifier}, which is a function of the true and generally unknown density ratio
between the generating distributions of signal and background in training data. In practice, good approximations can be obtained given enough data, flexible models, and a powerful learning rule. The previous relation is not unique for BCE-based probabilistic classification models, as it also holds for other approaches that minimise continuous relaxations of the zero-one loss and could be generalised for the multi-class case.

Viewing probabilistic classifiers as probabilistic density ratio estimators
allows us to abstract away from the specific machine learning techniques used to
construct the model ({\em e.g.} gradient boosting or neural networks trained by
stochastic gradient descent), and also provides a clear theoretical definition
for the best possible classifier, {\em i.e.} the one that optimally minimises
the risk or generalisation error of a classification problem, as a simple
function of the probability density ratios between the data-generating
distributions. Furthermore, density ratios can also be easily linked with the
statistical inference goals of data analysis and may effectively be used to
study the limitations of machine learning approaches.

We can also explore the previous formulation in the case when the generating
distributions of data are not fully specified, but depend on additional
unknown nuisance parameters $\boldsymbol{\theta}$. A classifier distinguishing samples from the data-generating
distributions $p(\boldsymbol{x} | \boldsymbol{\theta}, y=1 )$ and
$p(\boldsymbol{x} | \boldsymbol{\theta}, y=0)$ will still be approximating a
function of the density ratio
\begin{equation}
    r(\boldsymbol{x};\boldsymbol{\theta}) =
    \frac{
    p(\boldsymbol{x} | \boldsymbol{\theta}, y=1 )}{
    p(\boldsymbol{x} | \boldsymbol{\theta}, y=0 )}
\end{equation}
and hence will itself depend on the actual value of the parameters
$\boldsymbol{\theta}$. If we assume that the true value of the parameters is
fixed but unknown (which is the typical setting used for frequentist inference
in HEP), then the optimal classifier is not uniquely defined\footnote{In a
Bayesian setting, if the parameters are treated as random variables associated
with a prior probability density distribution $\pi(\boldsymbol{\theta})$, then
the optimal Bayes classifier can be uniquely defined, because parameter sampling
from $\pi(\boldsymbol{\theta})$ may be considered part of the data generating
procedure so the density ratio can be implicitly marginalised.}. For example, a
classifier trained using simulated data generated for specific parameters
$\boldsymbol{\theta}_\textrm{s}$ might not be optimal at classifying
experimental data observations that correspond to the unknown parameter value
$\boldsymbol{\theta}_\textrm{d}$. This is the main issue with nuisance
parameters from the perspective of the machine learning performance.

\subsection{Nuisance Parameters in Statistical Inference}

Another challenge with nuisance parameters, arguably the most relevant one, is
the way they affect our ability to extract useful information about our models
of nature from experimental data when carrying out statistical inference in the
form of hypothesis testing or interval estimation.
The effect of nuisance parameters is not solely a complication
for analyses based on machine learning approaches, as it also applies to
analysis based on manual variable transformations. However, as will be discussed
in detail in \sref{sec:inference-aware-approaches}, the presence of nuisance
parameters can put into question the role of supervised learning models in the
context of statistical inference, voiding them of the standard advantages that
otherwise make them so apt for dimensional reduction in physical analyses.

The previous concerns are closely related to the misalignment between the goal
of particle physics analyses --to infer information about our models of nature
given the data-- and the classification and regression objectives of supervised
learning approaches. For example, for inference problems based on mixture models where the mixture fraction is the only parameter of interest in the absence of
nuisance parameters, probabilistic classification models offer principled guarantees of optimality at the inference goal, as reviewed in \sref{sec:classification-not-enough}.
In general terms, the supervised learning task can be considered a proxy objective to obtain low-dimensional data transformations that are still informative about the
parameters of interest.

In this context, it is worth introducing the concept of \textit{summary
statistic}. For a set of $n$ experimental data independent and identically
distributed (i.i.d.) observations or events $D =
\{\boldsymbol{x}_0,...,\boldsymbol{x}_n\}$, where each $\boldsymbol{x} \in
\mathcal{X} \subseteq \mathbb{R}^d$ is a $d$-dimensional representation of the
event information at an arbitrary representational level ({\em e.g.} raw
detector readout, physical objects or a subset of columnar variables), a summary
statistic of the data $D$ is simply a function of the data that reduces their
dimensionality. An infinite number of different summary statistics can be
constructed for a given set of data, but we are generally only interested in
those which are as low-dimensional as possible conditional to approximately preserving the information relevant for the statistical inference goal of a given analysis. The low-dimensionality requirement is made necessary by the curse of dimensionality, due to $p(\boldsymbol{x}|\boldsymbol{\theta})$ not being known analytically and having to be estimated from a finite number of simulated observations. 

Most of the operations that reduce the dimensionality of the experimental data, either in terms of reducing the number of events (\textit{e.g.} trigger and event
selection) or its representation (\textit{e.g.} reconstruction, physical object
selection, feature engineering, multivariate methods, histograms) can be
viewed through the lens of summary statistics (see Chapter 3
of~\cite{DeCastroManzano:2701341}). Summary statistics used in high-energy physics
are thus often a composition of several transformations, yet for the purpose of
training machine-learning models we are usually interested the final components of the
type $\boldsymbol{s}(D) = \{ \ \boldsymbol{s}(\boldsymbol{x}_i) \ | \ \forall
\boldsymbol{x}_i \in D \}$ that are the product of the event-wise application of
a function
\begin{equation}
\boldsymbol{s}(\boldsymbol{x}) :
\mathcal{X} \subseteq \mathbb{R}^{d} \longrightarrow \mathcal{Y} \subseteq \mathbb{R}^{b}
\end{equation}
reducing the dimensionality of each event from the original feature space
$\mathcal{X} \subseteq \mathbb{R}^{d}$, which could be already a transformation
of the detector readout or set of engineered features, to a new low-dimensional
space $\mathcal{Y} \subseteq \mathbb{R}^{b}$.

Such a transformation may be used to reduce the data dimensionality from $n \times
d$ to $n\times b$. If $b$ is very small and because the observations are assumed
i.i.d., we could use simulated observations to estimate the probability density
$p(\boldsymbol{s}(\boldsymbol{x})| \boldsymbol{\theta})$ by non-parametric means
to carry out the inference goal. Most commonly in HEP applications, even simpler sample-wise statistics are instead constructed from $\boldsymbol{s}(D)$, such as the number of observations for which $\boldsymbol{s}(\boldsymbol{x})$ is within a given range (in so-called cut-and-count analyses) or simply a histogram of
$\boldsymbol{s}(\boldsymbol{x})$. For these simpler statistics, sets of simulations
of each process produced with different values of the nuisance parameters are
interpolated to model the effect of the nuisances, enabling the construction of likelihoods based on the product of Poisson terms.

The choice of the dimensionality reducing transformation determines the inference reach of a given analysis. Choosing a summary statistic is not easy
even in the absence of nuisance parameters, since naive choices of data transformation can very easily lead to a significant loss of useful information about the parameter of interest. Machine learning models, in particular probabilistic classification models trained to distinguish observations from different processes, are increasingly being used as an automated way to obtain
summary statistic transformations. This is because the output of probabilistic classifiers approximates density ratios  
: for simple hypothesis testing, density ratios are closely related to the optimal likelihood ratio test statistic, in the case of parameter estimation for the problem of inference of mixture fractions in the absence of nuisance parameters
\ifStandalone
.
\fi

It is of benefit for the ensuing discussion to succinctly recall how nuisance
parameters can be ``profiled away'' from a likelihood function in the extraction
of confidence intervals on parameters of interest; for a more general discussion
of how the effect of nuisance parameters is accounted for in physics
measurements see \textit{e.g.} \cite{Cranmer:2012sba,Cranmer:2015nia,Lista:2016chp}.


We consider the measurement of a physical quantity in statistical terms as a
problem of parameter estimation, whose solution relies on the specification of a
statistical model wherein those quantities appear as free parameters. Under the
assumption that experimental data conform to the specified model, the
measurement may be carried out by constructing suitable estimators for the
parameters of interest, which here we formulate through the specification of a
likelihood function. Letting $\theta$ identify the parameters of interest,
$\alpha$ describe systematic uncertainties affecting the model, and ${x_i}$,
$i=1...N$ be the collected data, understood to be a set of N random i.i.d.
variables, the joint probability density can be written as $p(x,\theta,
\alpha)$. This enables, {\em e.g.}, the construction of a likelihood function 
\begin{equation}
        \mathcal{L}(\theta, \alpha)=\prod_{i=1}^{N} p(x_i, \theta, \alpha).
\end{equation} 
If nuisance parameters $\alpha$ were absent in the above formulation, one would
proceed directly to construct estimators of the parameters of interest as 
\begin{equation}
        \hat{\theta} = \argmax_{\theta}  \mathcal{L}(\theta).
\end{equation}
The dependence on $\alpha$ can be dealt with by first obtaining the profile of
the likelihood function, $PL(\theta)$, by maximizing $\mathcal{L}$ as a function of the
nuisances,\par

\begin{equation}
        PL(\theta) = sup_{\alpha} \mathcal{L}(\theta,\alpha),
\end{equation}
and then proceeding as above. Uncertainties in the parameters of interest may
then be extracted from the curvature of the profile likelihood at its maximum
exactly as is done with $\mathcal{L}$ in the general case~\cite{Patefield1977-tu,
Cox1994-ah}. This ``profile likelihood method'' is conceptually simple and
practical to implement\footnote{In particle physics practice a widely used
implementation is the MINUIT package~\cite{James1975-nd}, which offers profile
likelihood evaluation through the MIGRAD routine.} if the likelihood is
differentiable with respect to the parameters, but may meet with technical
problems (\textit{e.g.} a slow convergence) as well as intractability for high
dimensional $\alpha$. The same issues affect in general the main alternative
solution, which consists in computing the marginalized
likelihood $\mathcal{L}_m(\theta)$ as
\begin{equation}
    \mathcal{L}_m(\theta) = \int \mathcal{L}(\theta, \alpha) p(\alpha) d\alpha.
\end{equation}
In both cases, knowledge of the PDF of nuisance parameters $p(\alpha)$ is mandatory for a meaningful solution. In a Bayesian construction $p$ may be a subjective prior for the nuisance parameters; in general it may be the result of any external constraint -{\em e.g.} an independent measurement. 
Whatever the source, any imprecision in $p(\alpha)$ will in general affect the parameter estimates with increases in bias and/or variance. 




\subsection {Toward Fully Sufficient Statistic Summaries} 

The reduction of {\em statistical} uncertainty on the estimate of parameters of
interest is a common goal of machine learning applications in experimental HEP.
A suitable summary statistic may be obtained by training a high-performance boosted decision tree or an artificial neural network on simulated sets of data.
The summary enables the extraction of the highest possible amount of information on the unknown true values of the physical quantities under measurement, {\em conditional on the validity of the underlying physics model} used to generate the training samples, as well as of specific assumptions on the value of relevant nuisance parameters. The crucial conditional clause above is usually hard to get rid of, because of the complexity of the problems, their high dimensionality, the typically unknown PDF of nuisance parameters, and/or the effects those parameters have on the physical model. When any one of the above effects play a role,  the obtained
summary statistic cannot be {\em sufficient}: being oblivious to a part of the feature space, it does not retain all the information contained in the data
relevant to the parameter estimation task, and is therefore liable to be
outperformed.

Notwithstanding the ubiquity of the above stated problem, the adjective {\em
optimal} is often employed when reporting physics analysis results,
usually in connection with incremental advances over state-of-the-art of the
employed techniques, for the common use case of classification performance in
signal versus background discrimination problems. The classical justification
for a claim that a chosen algorithm or architecture and its output (a
classification score) be {\em optimal} for the measurement task at hand is based
on examination of associated performance measures such as the integral of the
Receiver Output Characteristic curve (see  \cite{AIHEP:1.1}),
or on background
acceptance estimates at fixed purity --or {\em vice versa}. In the absence of
nuisance parameters those figures of merit are generally effective as a proxy to
classification performance, when their maximization closely tracks the
theoretical minimum value of the statistical uncertainty on the intermediate
physical parameter to which the classification algorithm is sensitive, such as,
{\em e.g.}, a signal fraction. Yet they are blind to the more general problem
connected with the subsequent extraction of, say, the cross section of a
physical reaction contributing to events labelled as signal, when uncertainties
of non-stochastic nature are included. 

\begin{figure}[!ht]
\includegraphics[width=\textwidth]{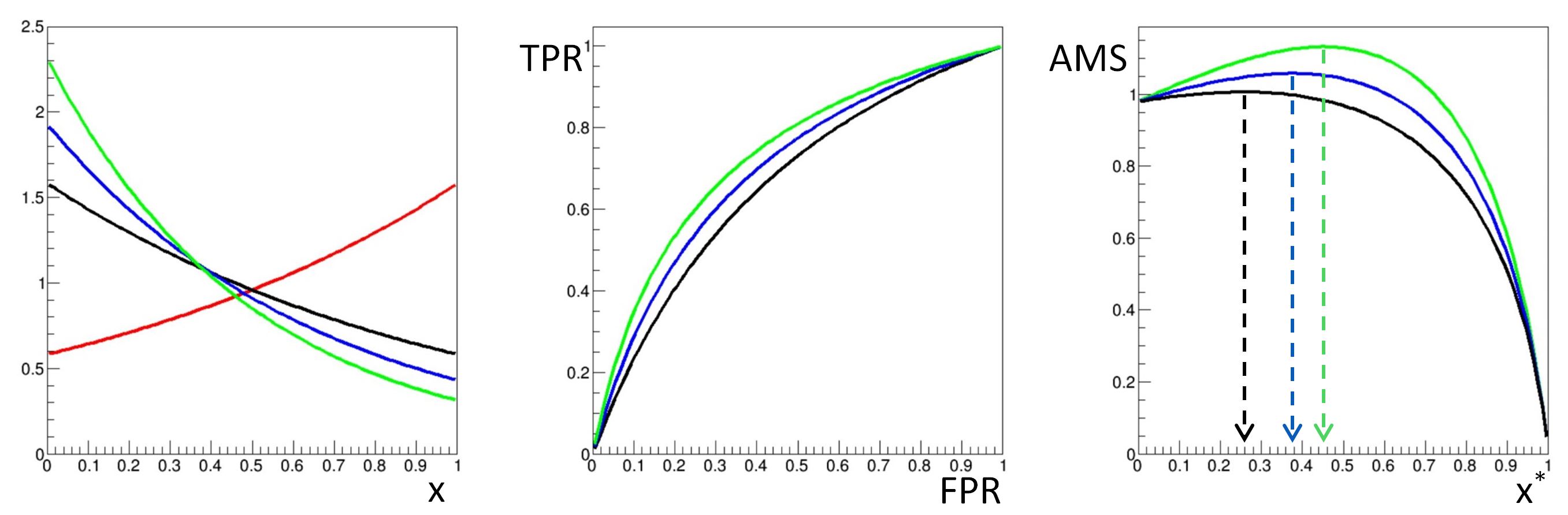}
\caption{\em A simple toy classification model. Left: the PDF of a signal
process (red) is compared to the PDF of background for three values of the
nuisance parameter, $\alpha=1.0, 1.5, 2.0$; center: ROC curves corresponding to
the three background distributions; right: values of the AMS figure of merit
(see text for details) as a function of the selection cut value $x^*$. The
location of maxima are shown by the corresponding arrows. }
\label{f:toymodel}
\end{figure}

A simple toy example may help pointing out the typical issues. Let us suppose
that a dataset includes events originated from a signal process of interest S,
in addition to ones coming from a known background source B. The typical output
of a classifier trained to distinguish the event classes may be the one shown in
Fig.\ref{f:toymodel} (left), which we parameterized using exponential functions
with normalized density functions in the $x \in [0,1]$ range:\par

\begin{equation}
S(x) = \frac{e^{x}}{e-1},
\end{equation}

\begin{equation}
B(x) = \frac{\alpha e^{-\alpha x}}{1-e^{-\alpha}}.
\end{equation}
Above, we have included a nuisance parameter $\alpha$ in the definition of the
background PDF. While the variations described by $\alpha$ in this example are very simple, the typical situation they mimic is a very common one, as the background distribution in HEP analysis problems is usually affected by significant uncertainties on its shape; $\alpha$ therefore describes what would be called a``background-shape systematic uncertainty". If we define the true positive rate (TPR) and false positive rate (FPR) of a selection criterion $x>x^*$ as\par

\begin{equation}
TPR(x^*) = \int_{x^*}^{1} S(x) dx = \frac{e-e^{x^*}}{e-1},
\end{equation}

\begin{equation}
FPR(x^*) = \int_{x^*}^{1} B(x) dx = \frac{e^{-\alpha x^*}-e^{-\alpha}}{1-e^{-\alpha}},
\end{equation}
then with simple algebra we may derive the ROC curve as the functional
dependence of TPR on FPR:\par

\begin{equation}
TPR(FPR) = \frac{e-[e^{-\alpha}+(1+e^{-\alpha})FPR]^{-1/\alpha}}{e-1}
\end{equation}

\noindent
By examining the shape of ROC curves resulting from different values of the
nuisance parameter $\alpha$ (see Fig.\ref{f:toymodel}, center), one may verify
the qualitative benefit of $B(x)$ densities peaking more sharply at $x=0$, which
correspond to larger values of $\alpha$. The performance of a classifier trained
under a given hypothesis for the nuisance parameter (say, $\alpha=1.5$) is then
liable to be under- or over-estimated if the value of $\alpha$ is uncertain; the
choice of a critical region $x>x^*$ corresponding to a pre-defined FPR will
similarly be affected, as will the resulting value of TPR.

In the given example the fraction of data selected in the critical region plays
the role of our summary statistic, as we have assumed that it is the only input
to a subsequent extraction of signal fraction. The fraction is of course
affected by the unknown value of the nuisance parameter $\alpha$, yet its value
alone does not retain information about it: the statistic is therefore not
sufficient. A sufficient statistic in this example would be the whole
distribution of the classifier output shown by the observed data; that choice
however fails to reduce, as desired, the dimensionality of the input, so it is not
an effective summary for the inference task.

To discuss how an optimal choice of $x^*$ based on the above densities may be
influenced by the presence of the nuisance parameter $\alpha$, we may consider
the figure of merit called \textit{approximate median significance}
(AMS)\cite{Cowan:2010js}, already introduced in \cite{AIHEP:1.1}:\par

\begin{equation}
AMS = \sqrt{2 \cdot [(N_s+N_b+N_r) \ln{(1+\frac{N_s}{N_b+N_r})} -N_s]}.
\end{equation}

\noindent
The AMS is a robust predictor of the statistical significance of an excess of
observed events if a signal of mean rate $N_s$ contributes to a data sample
assumed to be only composed of background events coming from a Poisson
distribution of known mean $N_b$; the regularization term $N_r$ reduces the
impact of Poisson fluctuations in low-event-counts situations, preventing
divergent behaviour when $N_b$ gets too low. If we set, {\em e.g.}, $N_r=10$ and
compute the AMS as a function of the selection cut $x^*$ for the three
considered values of $\alpha$ of our toy model, and for a choice of $N_s=20$,
$N_b=400$ in the data sample, we obtain the curves shown in the right panel of
Fig.\ref{f:toymodel}. It can then be observed, as expected, that the value of
$\alpha$ affects both the peak value of the figure of merit and the optimal
value of $x^*$ which achieves it.

The above toy model exemplifies how not only do nuisance parameters have the
power to modify the optimal working point of a ROC curve, but they also in
general affect the overall classification performance,  as well as the relative
merit of different classifiers. For that reason, standard supervised
classification techniques may not reach optimality unless they more broadly
address the conditionality issue stated above, or prove to be decoupled from
it\footnote{ It must be noted here that a possible misalignment between the
specification of the classification task and the true objective of the analysis
should always be considered. In the given example we studied the AMS score as a
robust proxy of the significance of an excess of signal events: such is a good
choice when the objective of the analysts is the discovery of a yet hypothetical
signal in the data. If, however their focus were rather the setting of the most
stringent upper limit on the signal rate --a common situation when the {\em a
priori} sensitivity of a search does not offer chances of a discovery-- then the
whole learning task and optimization criteria would have to be revised.}.

From a statistical point of view, in real-life situations ``all models are
wrong'', hence strictly speaking sufficient statistics that model the data may not exist! However, in most experimental situations approximate sufficiency is achievable, provided that the relevant nuisance parameters are included in the model and considered in the construction of the statistic. A number of brilliant ideas have been recently proposed to achieve that goal, in some cases effectively exploiting methods developed in Computer Science to endow learning algorithms of {\em
domain adaptation} capabilities --the flexibility to achieve good results on
data coming from one domain when trained with data coming from a different one,
such as {\em e.g.} the capability of driving a truck when trained to drive a
car. In the context of point estimation in experimental particle physics, the
different domains may involve a different relative importance of some of the
features, the absence of others, or imperfections in the
training model.

The growing interest in the development of new techniques to reduce or remove
the effect of nuisance parameters in physics inference, powered by the
availability of new machine learning tools and larger computing resources, has
brought the focus of experimentalists on the central problem of how to achieve a
true end-to-end optimization of physics measurements, and highlighted the need
to pay undivided attention to the expected total uncertainty on the parameters
of interest already in the training phase of learning algorithms. Below we
provide an overview of methods developed to address those issues, and discuss
their merits, applicability, and potential extensions.




\section{Nuisance-Parameterized Models}
\label{sec:nuisance-parameterized-models}




A direct way to account for the effect of nuisance parameters in the
construction of a summary statistic is to include them in the
physical model through a parameterization of their effect on the observable event features. This requires the injection in the model of knowledge of their PDF from an external prior, or from an ancillary measurement, and may or may not be practical to implement depending on the problem. 

In the simplest situations, as {\em e.g.} when the problem is low-dimensional, a
fully analytical solution may be sought. An example is offered by the
decorrelation of the ''N-subjettiness ratio'' variable $\tau_{21}$
\cite{Thaler:2011gf} designed in the context of searches for the two-body decay
of boosted resonances to reveal the resulting sub-structure in the produced
hadronic jets. The variable may be likened to a classification score as it
possesses large discrimination power against QCD background jets, but a
selection based on its value biases the distribution of reconstructed
''soft-drop'' mass successively used for inference, because of its dependence on
jet $p_T$, which can be here seen as a nuisance variable. As shown in
\cite{Dolen:2016kst}, an analytical parameterization of the dependence of
$\tau_{21}$ on jet $p_T$ removes almost entirely the distortions in the mass
spectrum. In the same context of boosted decay searches, similar results have
been obtained for the observable $D_2$ \cite{Moult:2017okx}.

In cases where experimental data are informative of the value of nuisance
parameters, one may try to exploit that dependence in the construction of
estimators for the parameters of interest. This situation was considered by
Neal\cite{Neal2008-om}, who addressed the question of how the extraction of a
sufficient summary for the signal fraction $\theta$ with a binary classifier is
affected by unknown parameters $\alpha$, when these modify the PDF of signal
$p_s(x,\alpha)$ and background events $p_b(x,\alpha)$. The likelihood for N
observations ${x_i}$,
\begin{equation}
    L(\theta,\alpha) = \prod_{i=1}^{N} \left[ \theta p_s(x_i,\alpha) + (1-\theta) p_b(x_i,\alpha) \right],
\end{equation}
may be rewritten as
\begin{equation}
    L(\theta,\alpha) = \left[ \prod_{i=1}^{N} p_b(x_i, \alpha) \right] \cdot \prod_{i=1}^{N} \left[ \theta \frac{p_s(x_i,\alpha)}{p_b(x_i,\alpha)} + (1-\theta)\right].
\end{equation}
The first term in the right-hand side of the last expression is not a constant
when nuisance parameters are present: factoring it out of the likelihood would
therefore cause loss of information, since background events alone carry
constraining power on the value of $\alpha$. The usual classifier task of
learning the ratio of signal and background PDFs $p_s(x)/p_b(x)$ is then no
longer sufficient to solve the problem as it would be if no nuisances were present.
The solution outlined in \cite{Neal2008-om} involves the construction of
low-dimensional summaries for both the nuisance parameters $\alpha$ and the
observable event features $x$, using {\em e.g.} a neural network. If good
parametric models of the summaries can be constructed one may use them for
inference, exploiting the informative power of the data themselves to constrain
nuisance parameters. Approximate sufficiency can in principle be obtained with
this recipe, if the parameterizations do not cause significant loss of
information.

In other cases of HEP interest no knowledge or constraints on a nuisance parameter may be
available, yet a parameterization of its effect on the observations successfully
solves the issue. The classical example of this situation is the search for a
new particle whose mass $M_{true}$ is unknown, when signal events exhibit smooth
variations in the momenta of the decay products as $M_{true}$ changes\footnote
{A dependence of the particle branching fractions on $M_{true}$ does not
complicate matters if the resulting acceptance variations are known.}. A
classifier trained to distinguish the new particle from backgrounds using signal
events simulated assuming a mass $M_1 = M_{true} + \alpha$ will consequently
suffers a progressive degradation in performance as $|\alpha|$ increases. This
was a common situation in early applications of binary classification to new
particle searches, which focused on a mass range of particular interest,
$M_{true} \simeq M_1$, and accepted the residual loss of power resulting for
$\alpha \neq 0$. A more performant, yet CPU-consuming,
solution\cite{Aaltonen:2012qt,Chatrchyan:2012tx} was to independently train a
set of classifiers $C_i$ using, in turn, data simulated assuming different mass
values ${M_i}$ for the unknown signal. This approach is still sub-optimal in a
general sense, since it does not fully exploit available resources (the
simulated data). Each classifier is ignorant about the information processed by
the other ones, as it only knows the precise mass hypothesis it corresponds to:
in general, it is not possible to interpolate the results of different
hypotheses. 

A way to avoid the above shortcomings, first proposed in \cite{Baldi:2016fzo},
is to parameterize the effect of the nuisance parameter in the construction of
the classifier. This may be achieved by including the unknown value of
$M_{true}$ within the set of features that describe simulated signal events; for
background events an arbitrary mass value, or one chosen at random for each
different training event, is correspondingly added. A suitable admixture of
training data with signal events corresponding to different $M_i$ hypotheses
spanning the range of interest may then be used in the learning
phase\footnote{Regardless of the {\em a priori} choice of signal admixture
employed in the training, the resulting inference cannot be considered Bayesian,
as the choice only affects the power of the classifier.}. The benefit of this
procedure is that it yields a interpolated classification score $C_k$ even for events
with mass values $M_k$ never seen during training. With this approach, a smoother
dependence is often expected if the classifier is a neural network rather than,
{\em e.g.}, a decision tree.

\begin{figure}[ht!]
\centerline{\includegraphics[width=0.8 \textwidth]{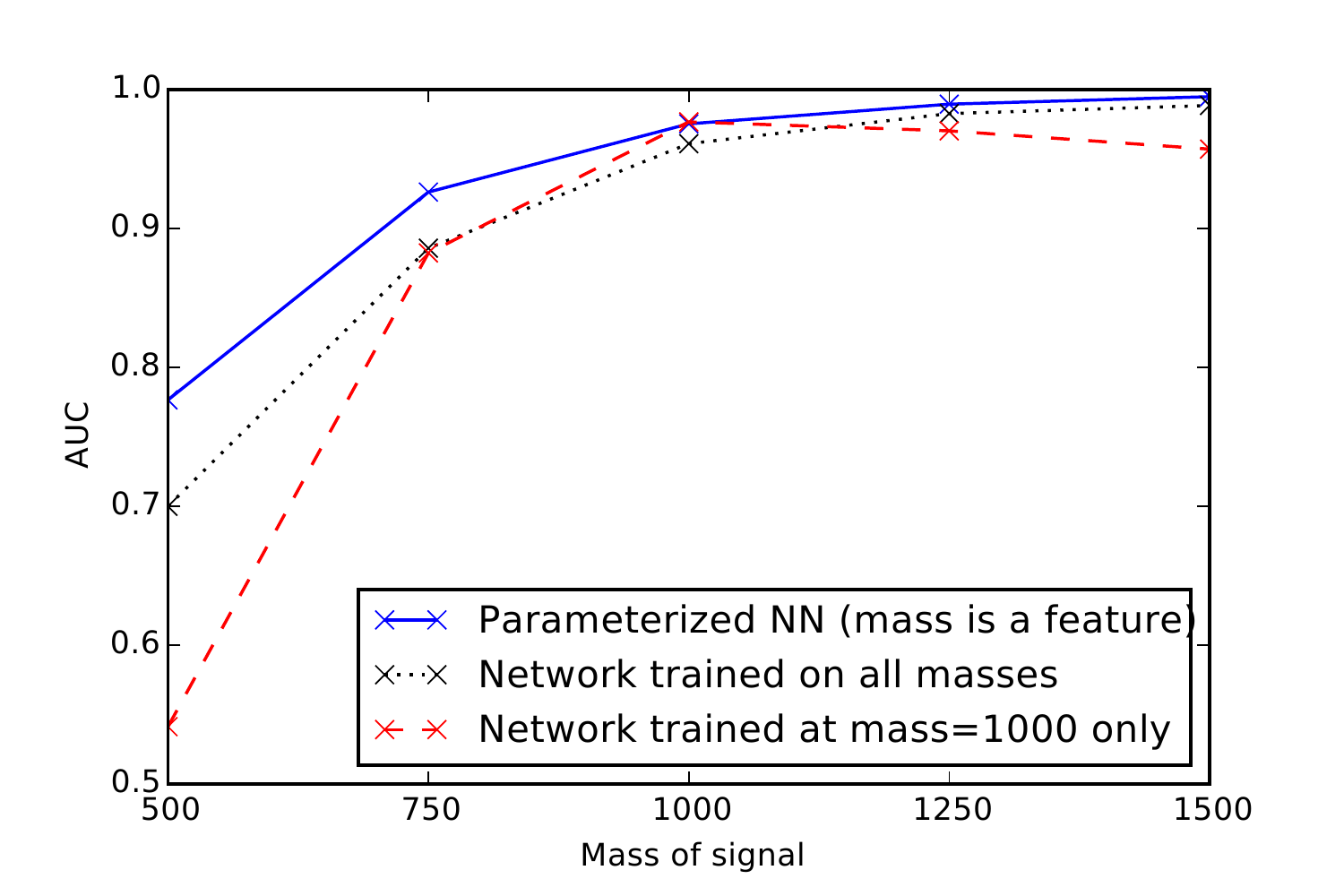}}
\caption{\em Area under the ROC curve for binary classification in the search
    for $X\to t \bar{t}$ on simulated ATLAS data as a function of the particle
    mass $M_i$ of test data. The parameterized NN (blue line) outperforms
    non-parameterized NN trained at a single mass value ($M_i=1000 GeV$, red
    dashed line) or trained with a mixture of signal samples for different $M_i$
    values (black dotted line). Reprinted with permission
    from\cite{Baldi:2016fzo}.}
\label{fig:baldifig8}
\end{figure}

The effectiveness of this strategy was demonstrated\cite{Baldi:2016fzo} using a
simple neural network architecture for the discrimination of a new particle $X$
decaying to top-antitop quark pairs from non-resonant $t \bar{t}$ backgrounds in
proton-proton collision data reconstructed by the ATLAS detector as simulated by
DELPHES\cite{deFavereau:2013fsa}. It was shown how for a given specific mass
hypothesis $M_1$ a parameterized NN performed similarly to a non-parameterized NN
trained with signal at the same mass, but it outperformed it for all other
masses (red curve in \fref{fig:baldifig8}), even when the non-parameterized NN
was trained with an admixture of mass values (black curve).

The parameterization of the dependence of observable variables on the latent
features of the underlying physical model --which include both interesting and
nuisance parameters-- has been more generally considered \cite{Cranmer2015-tq}
in the context of likelihood-free inference. The proposed algorithm performs a
dimensionality reduction of the data through a parameterization that is monotonic
with the likelihood ratio, allowing optimal inference via a calibration of the
output of a binary classifier. We refer the reader to
\sref{sec:likelihood-ratio-general} for more detail on this approach.



\section{Feature Decorrelation, Penalized Methods, and Adversary Losses}

When a direct parameterization of the effect of nuisance parameters on the
summary statistic used for classification proves ineffective or impractical to
implement, there are several possible alternatives. In a few specific
applications it proves sufficient to operate a suitable preprocessing of
training data that reduces or removes the dependence of the classifier output on
a variable sensitive to nuisance parameters. A second class of solutions aim to
make the classifier score insensitive to variations in the value of nuisances by
engineering a robust optimization objective for the classification task.
Finally, a more radical approach is to change the overall architecture of the
algorithm used in the search of the optimal solution, using adversarial
techniques to find the best compromise between signal discrimination and impact
of nuisances. Below we briefly discuss each of these approaches.

\subsection{Mass Decorrelation}

The intensive search for new physics carried out by the ATLAS and CMS
experiments in final states dominated by QCD backgrounds fostered, in the past
decade, the development of a number of imaginative new methods to increase signal
purity without modifying the shape of the distribution of reconstructed mass,
$M_{rec}$, of the hypothetical new particle, which is commonly used at the end of the selection step to estimate or limit the signal contamination in the sample.
Since the QCD background is complex to model reliably, a selection cut on the
output of a well-trained classifier does not guarantee optimal inference of the
presence of any signal, because the background retained by the cut is usually biased toward displaying a ''signal-like'' mass distribution. In this situation
$M_{rec}$ is not in itself a nuisance parameter; however, the reduction of its
discrimination power caused by the selection enhances the impact of
background normalization and shape uncertainties on the estimate of signal
fraction. Further, a reshaping of the background distribution complicates the
application of bump hunting techniques, {\em e.g.} by hindering the use of
data-driven background estimates based on mass sidebands.

The most straightforward way to reduce the dependence of a classification score
on $M_{rec}$ (or any other specific observable of relevance for inference
downstream of the selection) is called
''planing''\cite{Aguilar-Saavedra:2017rzt,Chang:2017kvc}. A simple way to
implement planing is to pre-select training samples for signal and background
such that they have the same marginal PDF in the variable one aims to
decorrelate, $p_S(M_{rec})^{sel} = p_B(M_{rec})^{sel}$. As the above corresponds
to making limited use of available training data, it proves more effective to
weight each event $i$ by a mass-dependent value $w(M_{rec,i})$ derived from the
PDFs of the two training datasets, $p_S(M_{rec})^{train}$ and
$p_B(M_{rec})^{train}$, \par

\begin{equation}
    w(M_{rec,i}) = \begin{Bmatrix}
             1/p_S(M_{rec,i})^{train}, & i \in S \\
             1/p_B(M_{rec,i})^{train}, & i \in B
          \end{Bmatrix}.
\end{equation}

\noindent
Weights, $w(M_{rec,i})$, enter directly the calculation of the loss function ({\em
e.g.} the binary cross-entropy) of the classifier in the training stage, but are
not used for validation and
testing. Planing has been shown to significantly reduce the dependence of
classifier output on the planed variable in specific situations, and due to the
simplicity of its implementation it may constitute a quite practical solution to
the problem; however, its effectiveness is limited when other event features in
one or both classes indirectly inform the classifier on the value of the planed
variable, if the latter --as is often the case-- carries discriminant power. 

In the context of searches for new physics in boosted hadronic jets, a
decorrelation of the output of a NN classifier from the mass of the boosted jet
was instead achieved by feature preprocessing based on principal component
analysis\cite{Aguilar-Saavedra:2017rzt}. The proposed method involved the PCA
rotation and standardization of 17 employed NN inputs (a basis set of
N-subjettiness variables $\tau_{N}^{\beta}$ 
proposed in \cite{Datta:2017rhs}) from trained data suitably binned in jet mass.
Besides avoiding the sculpting of the jet mass distribution of QCD background
events, the resulting classifier was shown to be effective for signal
discrimination also at signal masses for which it was not trained.

\subsection{Modified Boosting and Penalized Loss Methods}

As mentioned above, a decorrelation of the classifier output from a variable of
interest $x$ may be difficult to obtain with data preprocessing techniques when
other event features are informative of the value of $x$, especially if $x$
itself contains discriminant information. 

The search for new low-mass resonances in Dalitz plots\cite{Dalitz:1953cp} or
with amplitude analysis provides strong motivation to achieve uniformity of a
classifier selection as a function of kinematical variables of interest, as
systematic uncertainties may be greatly amplified by the unevenness of selection
efficiency. The first algorithm explicitly targeting that use case is {\em
uBoost}\cite{Stevens:2013dya}, which relies on boosted decision trees to improve
signal purity. The method builds on the standard AdaBoost
prescription\cite{Freund1997} of increasing the weight of training events
misclassified by the decision tree built in the previous iteration of the BDT
sequence, augmenting it by modifying the weight of signal events depending on
the disuniformity of the selection. If $w_i^{n-1}$ is the weight of event $i$ at
boosting iteration $n-1$, the new weight is computed as\par
\begin{equation}
        w_i^n = c_i^n u_i^n w_i^{n-1} 
\end{equation}
\noindent
where $c_i^n= exp(-\gamma_i p_i^{n-1})$ is the AdaBoost classification weight,
with $\gamma_i=+1$ $(-1)$ for signal (background) events and where $p_i$ is the
prediction of the previous decision tree in the series.

The uniformity weight $u_i^n$ is defined as the inverse of the density of signal
in the proximity of event $i$, and is computed with the k-nearest-neighbor
algorithm; for background events $u_i$ is set to unity.
Since it is necessary to consider many different values of signal efficiency
in the construction of
the final BDT score and to the use of kNN, the CPU cost of training with uBoost
is higher than that of a regular BDT, but not prohibitive in practical
applications. Tested on a Dalitz analysis, the method was shown to achieve the
wanted uniformity with almost no loss in classification
performance\cite{Stevens:2013dya}.

Following on the thread of uBoost, a number of interesting alternatives to
achieve uniform selection efficiency of a BDT classifier were introduced in
\cite{Rogozhnikov:2014zea}, again targeting the use case of Dalitz plot
analysis. The algorithm called kNNAdaBoost achieves the uniformity
goal by modifying the AdaBoost weights to include information on the
classification probability of k nearest neighbors to each event,\par

\begin{equation}
        w_i^n = w_i^{n-1} exp \left[ -\gamma_i \sum_j a_{ij} p_j \right],
\end{equation}

\noindent
where the $a_{ij}$ matrix collects information on the density of events of the
same class around event $i$, by setting $a_{ij} =1/k$ if $j$ is among the k
neighbours of $i$ and $=0$ otherwise. Other methods proposed in
\cite{Rogozhnikov:2014zea} involve the use of $a_{ij}$ in the loss of the
classifier, minimized with the use of gradient boosting. These techniques are
shown to improve over uBoost by achieving better uniformity in specific use
cases.

More recently, the issue of the decorrelation from variables of interest or,
more generally, robustness to nuisance parameters has been addressed by using
neural network classifiers, adding suitable regularizer terms to their loss
function. An option discussed in \cite{Kasieczka:2020yyl} is to use, for that
purpose, a measure of the extent to which two sets of features $\vec{x}$,
$\vec{y}$ are independent \footnote{ In their work Kasieczka and Shih discuss
the classical single-dimensional case when $x=M_{rec}$ is the mass of a searched
particle and $y=c$ the output of the classifier itself, but the extension to
multi-dimensional problems is straightforward.}. The proposed measure is dubbed
{\em DisCo} (''distance correlation'')\cite{Kasieczka:2020yyl}, a function of
the considered features which can be constructed by first defining a distance
covariance\par

\begin{equation}
        dCov^2(X,Y) = \langle |X-X'||Y-Y'|\rangle + \langle |X-X'| \rangle \langle |Y-Y'| \rangle - 2 \langle |X-X'||Y-Y''| \rangle
\end{equation}

\noindent 
where $| \cdot |$ is the Euclidean vector norm and $(X,Y)$, $(X',Y')$ and
$(X'',Y'')$ are {\em i.i.d.} pairs from the joint distribution of the two
features; brackets indicate taking averages. The distance correlation, defined as \par

\begin{equation}
        dCorr^2(X,Y) = \frac{dCov^2(X,Y)}{dCov(X,X) dCov(Y,Y)}
\end{equation}

\noindent
is then bound between 0 and 1, and is null only if $x$ and $y$ are fully
independent. $dCorr^2(X,Y)$ is differentiable and can be computed from batches of data
samples; its value can be profitably added as a penalty term to the loss of the
classifier, once multiplied by a positive hyperparameter $\lambda$ controlling
its strength. The multiplier allows to gain control over the acceptable amount
of interdependence of $x$ and $y$ achieved by a minimization of the penalized
loss. In the single-dimensional application considered
in\cite{Kasieczka:2020yyl} DisCo proves competitive or advantageous over, {\em
e.g.}, adversarial setups (see below) or other methods; further studies are
needed to gauge its performance in more complex situations. 

A similar approach is taken \cite{Wunsch2019-za} in a study more explicitly
aiming at a reduction of the dependence of classifier score from nuisance
parameters $\alpha$. 
In the proposed technique the $n$-bin histogrammed distribution ${\cal{N}}_k$ of classifier output $f(x)$ from input features $x$ is first made differentiable with the use of a Gaussian smoothing with functions ${\cal{G}}_k$, \par

\begin{equation}
        {\cal{N}}_k(f(x)) = \sum_b {\cal{G}}_k(f(x))
\end{equation}

\noindent
where $k$ runs on the bins and $b$ runs on the training events in a batch. The
usual loss $L_0$ of the classifier can then be penalized by a term derived from
the difference in smoothed bin counts of the original output $f(x)$ and its
nuisance-varied value $f(x+\alpha)$, \par

\begin{equation}
        L(\lambda) = L_0 + \lambda \frac{1}{n} \sum_k \left(\frac{{\cal{N}}_k(f(x))-{\cal{N}}_k(f(x+\alpha))}{{\cal{N}}_k(f(x))}\right)^2.
\end{equation}

\noindent
The modified loss effectively decouples the classifier output from the value of
$\alpha$, both in a synthetic example and in the benchmark problem of $H \to
\tau \tau$ discrimination proposed in the ATLAS kaggle
challenge\cite{Adam-Bourdarios:2015pye}, where the $\tau$ lepton momentum scale is
considered as the nuisance parameter.

\subsection{Adversarial Setups}

The construction of an adversarial setup where two independent neural networks
are pitched one against the other in the search for the optimal working point in
a constrained classification problem may be considered an extension, if not the
logical next step, of the penalized loss methods discussed above. In fact, the
global loss function is still the combination of two parts, one of which is the
usual classification loss ({\em e.g.} a Binary Cross-Entropy term) and the other is a
penalization contributed by the adversary, usually modulated by a regularization
multiplier $\lambda$. The difference is that adversarial architectures create a
conceptual symmetry between the classification task aiming at a
signal-background separation and the discrimination of different values of a
nuisance parameter, putting the two minimization problems on equal footing. 

The idea of using a classifier trained to discriminate between data from different domains to bound the error of a binary classifier trained in one domain and applied to a different domain dates back to early computer science research\cite{NIPS2006_2983,
BenDavid2009}. These seminal works are foundational in the more modern domain adversarial approaches which formulate the domain adaptation approach through a min-max learning objective.
Applications in HEP arise when training and test data come from different domains (a source and a target one), or when training data are simulated by an imperfect
model of real (test) data. It was shown that robust classification can be
achieved in such situations if one can find a suitable representation of the
data which is maximally insensitive to their source. An adversarial neural
network is thus tasked to learn such a representation while competing with one
that tries to achieve maximal separation of labelled classes of training
data\cite{ajakan2014domain}.


The first proposal of adversarial neural networks to achieve robustness to
systematic uncertainties in HEP problems was the one of Louppe, Kagan and
Cranmer\cite{NIPS2017_6699}, who showed the feasibility of using adversarial
techniques to make the classification score $f(X; \theta_f)$ a "pivotal
quantity" in the statistical sense \cite{Degroot1977ProbabilityAS}, {\em i.e.}
one whose distribution is independent on the value of nuisance parameters $z$;
above, $\theta_f$ are the parameters of the classifier, and $X$ denote the data.
If one further denotes the adversary,$r$, with parameters $\theta_r$, whose
task is to discern values of $z$ from the output value $f(X,\theta_f)$ of the
classifier, the loss functions of the two networks may be succinctly written
${\cal{L}}_f(\theta_f)$ and ${\cal{L}}_r(\theta_f,\theta_r)$, and a simultaneous
training can be carried out by using the value function\par
\begin{equation}
        E(\theta_f, \theta_r) = {\cal{L}}_f(\theta_f) - {\cal{L}}_r (\theta_f,\theta_r)
\label{eq:adversaryloss}
\end{equation}

\noindent
which can be optimized by the minimax solution\par
\begin{equation}
        \hat{\theta_f}, \hat{\theta_r} = arg min_{\theta_f} max_{\theta_r} E(\theta_f,\theta_r).
\end{equation}

\begin{figure}[!ht]
 \centerline{\includegraphics[width=0.8\textwidth]{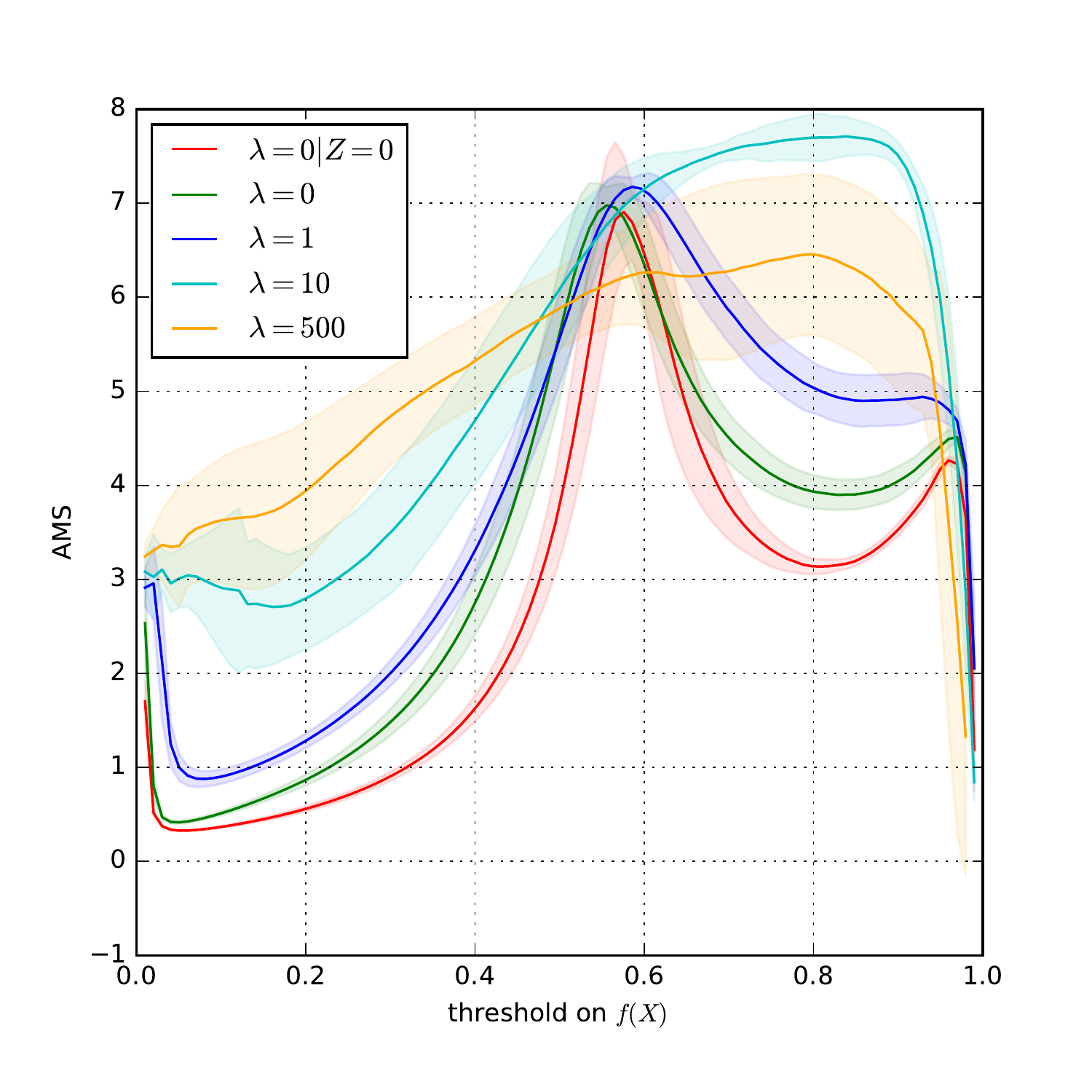}}
 \caption{\em AMS score as a function of classifier score for a binary classification task, for different values of the hyperparameter $\lambda$ modulating the loss penalization, and for the case when no nuisance parameter is present. For $\lambda=10$ an advantageous tradeoff of classification accuracy and robustness to the nuisance is obtained at high classification scores. Reprinted with permission from \cite{NIPS2017_6699}.} 
 \label{f:amslouppe}
 \end{figure}

Convergence to the optimal solution cannot be guaranteed if the nuisance
parameters shape the decision boundary directly. In that case a hyperparameter
$\lambda$ multiplying the adversary loss ${\cal{L}}_r$ may be introduced in
\eref{eq:adversaryloss}, and a search for approximate optimality must be
performed. As an example, Louppe \textit{et al.} consider both a
synthetic example and a HEP use case when the nuisance parameter $Z$ is
categorical, describing the absence ($Z=0$) or presence ($Z=1$) of pile-up in
LHC collisions data. In the latter case they show (see Fig.\ref{f:amslouppe})
how an effective compromise between the classification and the pivotal tasks may
be found by a tuning of $\lambda$.

The application of the above technique to the discrimination of the decay of
boosted heavy particles in a situation where background systematics affect the
inferential step downstream of the NN-based selection was considered in
\cite{Shimmin:2017mfk}. In their work, authors showed how the relevant utility
function in the problem --the significance of a resonant signal in the data,
once systematic uncertainties were accounted for-- was indeed maximized by an
adversarially trained classifier, despite its slight degradation of separation
power with respect to a non-adversarial classifier. 

A further comparison of the effectiveness of the adversarial training proposed
in \cite{NIPS2017_6699} to alternatives based on data augmentation and tangent
propagation, for the goal of optimizing classification in presence of nuisance
parameters, was produced in \cite{Estrade2017AdversarialLT}. The considered HEP
problem was the one of $H \to \tau \tau$ discrimination from backgrounds
proposed by the Higgs Kaggle Challenge\cite{Adam-Bourdarios:2015pye}, where an
uncertainty on the $\tau$ lepton energy scale was introduced and propagated to
the input features of signal and background. Besides a baseline,
non-systematics-aware neural network classifier, they employed in their
comparison a data augmentation method based on training datasets constructed so
as to appropriately sample the relevant range of values of the nuisance
parameter. Finally, the tangent propagation method consisted in modelling
nuisance parameters as coherent geometric transforms of the event features,
operated by differentiable functions; a regularization of the model was provided
through the derivative of the classifier score on the nuisance parameter value,
as introduced by \cite{NIPS1991_536}. The comparison showed that adversarial
learning had a minor advantage over data augmentation, although further work was
deemed necessary to achieve more conclusive results on the matter. Tangent
propagation was instead shown to be unsuccessful on the specific problem
considered, due to large uncertainties introduced in the geometrical
transformation caused by the large class overlap in the feature space.

We conclude this survey of applications of adversarial techniques to constrain
the effect of nuisance parameters with a mention of two very recent studies. The
first, by Blance, Spannowsky and Waite \cite{Blance:2019ibf}, examines
adversarial classification as a preliminary step to the use of autoencoders for
unsupervised classification, to verify their effectiveness in reducing the
dependence of the autoencoder task on systematic uncertainties. They apply this
idea to the search of resonances decaying to semileptonic $t \bar t$ final
states, showing promising independence of the resulting classification task on
the considered smearings of the input models. A second interesting recent study
\cite{Englert:2018cfo} attacks the problem of theoretical uncertainties with
adversarial networks. As uncertain theory parameters affect the data in a
coherent way, they can be controlled more effectively than experimental ones in
machine learning applications. The authors consider the case of searches for new
physics in events with a Higgs boson and a high-momentum jet, where
renormalization and factorization scale variations heavily affect the
predictions of standard model backgrounds, making traditional discrimination
methods unreliable. Sensitivity to new physics can be retained by an adversarial
technique which ensures robustness to theoretical scale uncertainties, with
however smaller, and more realistic, discrimination power.





Overall, adversarial methods discussed in this section prove effective to
achieve approximate independence of the classification from the value of
selected input features. In general, however, there is no guarantee that the
resulting equilibrium point between the two competing tasks be optimal for the
final goal of the analysis in which they are embedded. For this reason, the
hyperparameter $\lambda$ governing the tradeoff between the two losses must be
optimized independently. More direct ways to strive for a complete optimization
of classification in physics measurements and searches are examined in 
Sec. 5.



\section{Semi-Supervised Approaches}


\ifStandalone
In Sec. 1
\fi
nuisance parameters were introduced as
additional parameters that account for the limitations of the description of the
data and have to be accounted for making accurate statistical statements. Given
that most machine learning models in HEP are usually trained using simulated
observations, the resulting models could only aspire to be optimal at the task at hand, typically classification or regression, for the specific configuration of nuisance parameters used for data generation. The previous sections discussed some solutions to this problem such as parameterizing the model or decorrelating its output using additional loss terms. In this section, we review alternative approaches that are based on using actual experimental data to complement or substitute simulated samples in the model training procedure, focusing on how these techniques could help to deal with nuisance parameters.

Experimental data are the source of information used to test hypothesis or estimate parameters given a model. Models are usually based on detailed simulations of the underlying physical processes and the detector response, providing, in general, a quite good although not perfect description of the data. Oftentimes, experimental data from well-known processes are also used to cross-check the accuracy of the description by the model and to estimate correction factors and associated uncertainties as necessary. These calibration procedures, which also constitute statistical inference analyses in their own right, provide a mechanism to improve possible mismodelling issues and obtain data-based estimates for nuisance parameter constraints. While general calibrations are typically performed experiment-wide, more detailed calibrations are often carried out for specific analysis scenarios to improve their precision and discovery reach, for example using an independent subset of data that is expected to be well-modelled to further correct or constrain known unknowns at the inference stage. In some cases, yet arguably not often in analyses that use machine learning to reduce the dimensionality of their summary statistics, known properties of experimental data allow us to use a well-understood subset to model one of the mixture components.

The interrelation between experimental data and the generative model and its
parameters in HEP is thus more involved than its ideal depiction in statistical literature. When training supervised machine learning models using simulated data, the expected performance at the objective task in experimental data would benefit from training and validation datasets that are well-calibrated and correspond to the best estimates of the parameters of interest\footnote{For completeness, we note that even when the machine learning model is not trained with the most accurate description of the data, it is still possible to make calibrated
statistical statements, as long as known unknowns are properly accounted for in the statistical model used for inference procedure.}. Leaving aside issues with whether the supervised learning task is a good proxy of the analysis inference goal when nuisance parameters are important, 
which will be discussed in
\sref{sec:inference-aware-approaches}, 
we review here methodologies that use experimental data during training to close the gap between the performance at the inference task between real and simulated data.

Many of the efforts to achieve the above goal are based on innovations from \textit{weak supervision} and \textit{semi-supervised learning}, that focus on the problem of learning useful models from partial, non-standard or noisy label information. In this context, when considering a classification task, simulated
observations can be considered as fully labelled data that provide a possibly
imperfect description, while real data observations can be thought as
unlabelled or very sparsely labelled mixtures from different classes, which however do not suffer from the same imperfection. For example, Dery \textit{et al.} \cite{Dery2017-zm} proposed an approach based on \textit{learning
from label proportions} (LLP), where instead of a label per observation a neural
network is trained only based on the class proportions for a given set are known
in average using a custom loss. They validate the method on a quark versus gluon
tagging example problem, finding that it can be used to obtain a similar
performance to that of a fully supervised classifier, while being more robust to
simulation mismodelling of the input variables.

\begin{figure}
\centerline{\includegraphics[width=0.7\linewidth]{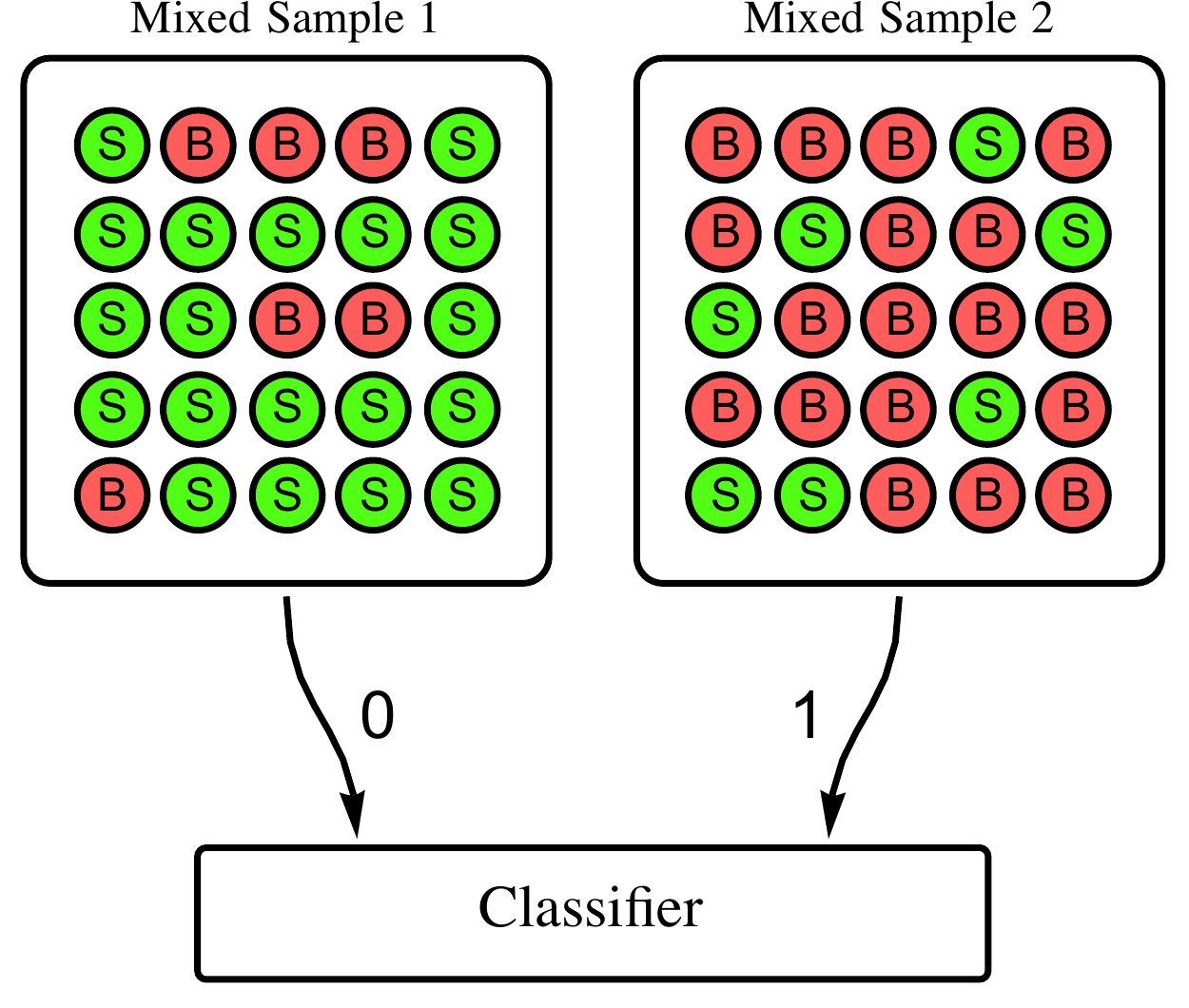}}
\caption{An illustration of the CWoLa framework. Figure and description by the
authors of \textsc{CWoLA}~\cite{Metodiev2017-kh} and licensed under CC BY 4.0.}
\label{fig:framework-cwola}
\end{figure}

One of the potential advantages of approaches based on learning from label
proportions (and weak supervision more generally) is that in principle they
could be extended to train the classifier directly using data from the experiment.
However, the
previous approach based on LLP requires at least knowledge of the label
proportions in the mixed samples, which might not be known at training time. To address this limitation, Metodiev \textit{et al.}~\cite{Metodiev2017-kh} proposed a new paradigm referred to as classification without labels (\textsc{CWoLA}), where the
classifier is trained to distinguish between two mixed sample with different
(and possibly unknown) component fractions. This also simplifies the previous
approach because it is based on standard classification loss, where the label
is not the observation class but an identifier of the mixed sample it belongs
to, as depicted in \fref{fig:framework-cwola}. The authors prove that the
optimal binary classifier (in the Bayes sense) for distinguishing samples from
each of the mixed samples is a simple function of the density ratio
between the components. Furthermore, they demonstrate that \textsc{CWoLA} as
well as LLP work similarly to a fully supervised classifier on pure samples,
using practical examples such as a quark/gluon discrimination problem.

While \textsc{CWoLA} has a wider range of applicability than standard LLP techniques, it also requires two (possibly smaller) mixed test data samples with known fractions to establish operating points. After its introduction, two other studies have applied variations of \textsc{CWoLa} to sample use cases.
Cohen \textit{et al.}~\cite{Cohen2018-ph} applied weakly supervised neural networks to
the new-physics search of gluino production using fast simulation samples, and
also demonstrated that weak supervision can perform similarly to full
supervision and that it is robust to certain types of mismodelling. Further work
by Komiske \textit{et al.}~\cite{Komiske2018-wp} has shown that weak
supervision approaches scale well to problems with high-dimensional inputs and
larger models, by successfully applying LLP and \textsc{CWoLa} to the
quark/gluon discrimination problem directly using a convolutional network model
applied directly to jet images.

Nevertheless, probably the main advantage of weak supervision techniques such as
LLP and \textsc{CWoLa} is that, in principle, they could allow the use of real
data during the training procedure. The use of experimental data
for training with this family of techniques has not been demonstrated in HEP
practice so far. In the best case scenario, a weakly supervised classifier trained with data could be used to extract the optimal classifier (in the Bayes sense) between each of the mixture components (e.g. signal and background). The output of this classifier could then be used to select or construct a summary statistic to carry out the inference goal of the analysis. Yet most likely, the model would have to be constructed using simulated observations that are subjected to the effect of nuisance parameters. Hence a potentially Bayes-optimal weakly supervised classifier would suffer the same pitfalls as any other classifier in relation with the
analysis inference goal, as we discuss below, in
\ifStandalone
Sec. 5.
\fi

Furthermore, if experimental data are used during training, then the model might be overfitted to the particular statistical fluctuations of the dataset, so an experimental data splitting scheme, or the use of experimental data from an independent subset, might be needed to avoid biased estimations. If the data representing different mixture fractions are taken from different control regions, the previous caveat could be avoided but then the density of the components for each of the mixed samples is not the same, so that the basic theoretical assumption of \textsc{CWoLa} or LLP does not hold. The fundamental assumption also does not hold if the distributions in the control region exhibit different correlations than in the signal region.

In conclusion, while weak supervision could be useful to build classifiers that might benefit the model classification performance, due to their being more robust to certain types of mismodelling, existing practical approaches do not fully address the issue of dealing with nuisance parameters.

\section{Inference-Aware Approaches} \label{sec:inference-aware-approaches}


The approaches discussed so far use diverse methodologies in order to overcome
situations where the data generating process is not perfectly known and thus
the performance of the supervised learning task considered (typically classification)
might be degraded once it is applied on real data. However, recent work has shown
that some of the innovations in the field of machine learning are flexible
enough so as to be re-purposed to deal more closely with the statistical
inference objective of HEP analyses.

The solutions discussed in this section thus move away from the overall goal of
optimising models to become performant at proxy supervised learning tasks such
as classification, and attempt to frame the problem directly as one of
statistical inference. This change of paradigm is often referred to as
likelihood-free or simulation-based inference, and is a rapidly evolving line of
research, with applications within particle physics as well as in other
scientific domains that heavily rely on complex generative models, such as
epidemiology or cosmology.

For a broader overview of the techniques proposed to deal with this problem and
their role in particle physics we refer to 
\ifStandalone
other
general reviews~\cite{Cranmer2019-xo}.
\fi In this section we instead focus on how some of these inference-aware approaches could be useful to deal with nuisance parameters in the context of particle physics. Given that most of these solutions already cast the problem in the form of statistical inference on a set of parameters given the data, it is not surprising that nuisance parameters could be incorporated or dealt with in a principled way for many of these methods.

\subsection{Why are classification and regression not enough?}
\label{sec:classification-not-enough}

Before delving into these new techniques, it is worth considering the limitations
of classification and regression as proxy supervised tasks from the point of
view of statistical inference. For simplicity, let us consider the paradigmatic
problem of inference about the mixture coefficient in a two-component mixture
model, which can be thought of as the basis for both cross section measurements and new physics searches:
\begin{equation}
    p(\boldsymbol{x}| \mu, \boldsymbol{\theta} ) =
    (1-\mu) p_b(\boldsymbol{x} | \boldsymbol{\theta})
    + \mu p_s(\boldsymbol{x} | \boldsymbol{\theta})
\label{eq:mixture_structure}
\end{equation}
where $\mu$ is a parameter corresponding to the signal mixture fraction,
$\boldsymbol{x}$ is the event feature space and $\boldsymbol{\theta}$ are other
parameters which the component distribution functions might depend on. For the
problems of relevance for machine learning techniques, we may assume that the
component probability density functions for signal $p_s(\boldsymbol{x} |
\boldsymbol{\theta})$ and background $p_b(\boldsymbol{x} | \boldsymbol{\theta})$
are not known parametrically, yet we have access to random samples from a
simulator that is able to model them implicitly.

The relation between the density ratio approximations from
\eref{eq:density_ratio_clf} and the typical problems of inference in HEP may be studied using two different statistical constructions: likelihood ratios or summary statistics. Both approaches lead to equivalent conclusions regarding the limitations of classification as a means of obtaining useful transformations for statistical inference in the presence of nuisance parameters, but they are both relevant because they imply ways of framing the problem which map very well to different families of new techniques built to address this issue which we discuss later in this section.

Let us start with likelihood ratios, which can be generally defined for a set of
$n$ data observations $ D = \{\boldsymbol{x}_0,...,\boldsymbol{x}_n\}$ between
two simple hypotheses $H_0$ and $H_1$ as:

\begin{equation}
    \Lambda( \mathcal{D}; H_0, H_1) =
    \frac{p(D| H_0)}{p(D| H_1)} =
    \prod_{\boldsymbol{x} \in \mathcal{D}}
    \frac{p(\boldsymbol{x}| H_0)}{ p(\boldsymbol{x} |H_1)}
\label{eq:likelihood_ratio}
\end{equation}
where the last expansion requires independence between observations, and where we note that the quantity $p(\boldsymbol{x}| H_0) / p(\boldsymbol{x} |H_1)$ is a density ratio and could be approximated as discussed before by training a probabilistic classifier to distinguish samples generated under each hypothesis. From the Neyman-Pearson lemma~\cite{NeymanPearson1933}, we know the likelihood ratio is the most powerful test statistic to distinguish the two simple hypotheses $H_0$ and $H_1$ at given significance level $\alpha=P(\Lambda(D; H_0, H_1) \leq
t_\textrm{cut})$, for any threshold $t_\textrm{cut}$.

Going back to problems where hypotheses have a mixture structure like the one
discussed in \eref{eq:mixture_structure} and differ in their mixture
composition, this would mean training a classifier between samples generated
from $p(\boldsymbol{x}| \mu, \boldsymbol{\theta})$ for the specific mixture
fractions $\mu_0$ and $\mu_1$ that characterise each of the hypothesis
$p(\boldsymbol{x}| H_0)=p(\boldsymbol{x}| \mu_0, \boldsymbol{\theta})$ and
$p(\boldsymbol{x}| H_1)=p(\boldsymbol{x}| \mu_1, \boldsymbol{\theta})$, which
would become rapidly cumbersome if we are dealing with multiple tests for a set of different $\mu_0$ and $\mu_1$ values. Luckily, each factor in
the likelihood ratio from \eref{eq:likelihood_ratio} can be expressed in the
following manner:
\begin{eqnarray}
    & & \frac{p(\boldsymbol{x}| H_0)}{ p(\boldsymbol{x} |H_1)} = \frac{(1-\mu_0)
    p_b(\boldsymbol{x} | \boldsymbol{\theta}) + \mu_0 p_s(\boldsymbol{x} |
    \boldsymbol{\theta})}{(1-\mu_1) p_b(\boldsymbol{x} | \boldsymbol{\theta}) +
    \mu_1 p_s(\boldsymbol{x} | \boldsymbol{\theta})} \\
    &=& \left (
    \frac{1-\mu_1}{1-\mu_0} + \frac{\mu_1}{1-\mu_0}
    \frac{p_s(\boldsymbol{x} | \boldsymbol{\theta})}{p_b(\boldsymbol{x} | \boldsymbol{\theta})}
    \right )^\textsuperscript{-1} + \bigg(\frac{1-\mu_1}{\mu_0} \left (
    \frac{p_s(\boldsymbol{x} | \boldsymbol{\theta})}{p_b(\boldsymbol{x} | \boldsymbol{\theta})}
    \right )^\textsuperscript{-1} + \frac{\mu_1}{\mu_0}
    \bigg)^\textsuperscript{-1}
\label{eq:likelihood_ratio_expansion}
\end{eqnarray}
so for a given pair $\mu_0$ and $\mu_1$ the density ratio between hypotheses
factor in the likelihood ratio is a bijective function of the ratio
$p_s(\boldsymbol{x} | \boldsymbol{\theta})/p_b(\boldsymbol{x} |
\boldsymbol{\theta})$. That quantity can be approximated by training a
probabilistic classifier to distinguish signal and background simulated samples,
which is computationally more efficient and easier to interpret intuitively than
directly the ratio $p(\boldsymbol{x}| H_0) / p(\boldsymbol{x} |H_1)$ mentioned earlier.

A likelihood ratio approximation can thus be obtained in the case of two simple
two-component mixture hypotheses that only differ in the mixture fractions by
plugging the output of a probabilistic classifier $c(\boldsymbol{x})$ trained to
distinguish signal and background observations in
\eref{eq:likelihood_ratio_expansion} with the corresponding values of $\mu_0$
and $\mu_1$ in \eref{eq:likelihood_ratio}. Oftentimes, the calculation
of the likelihood ratio is not necessary
because the classifier output directly contains all the relevant information
about the ratio approximation. Hence the classifier output
can be used directly as a summary for
inference with the help of histograms or non-parametric density estimation
techniques, with the added advantage that is typically a $[0,1]$ bounded variable
and thus easy to interpret. It is worth mentioning that the relation between the
likelihood ratio and the density ratios of the pair of mixture components
can also be useful in the
multi-component mixture setting. In that case, the likelihood ratio factor can be expressed in
terms of the density ratios that can be obtained for each pairwise component
classification problems~\cite{Cranmer2015-tq}.

Within this framework, the usefulness of probabilistic classifiers that
distinguish signal and background observations is that they can be used to
approximate the likelihood ratio, which is the most powerful summary statistic
for two simple hypothesis that differ only on the the mixture fraction
parameters. If the hypotheses are not fully specified, i.e. they depend on
additional parameters (the dependence with the mixture fractions can be factored
out as discussed before), the likelihood ratio as defined in
\eref{eq:likelihood_ratio} also depends on these parameters. The
Neyman-Pearson lemma does not hold when parameters are varied nor for composite
generalisations such as the profile likelihood ratio. Hence, when nuisance
parameters are important, a fixed probabilistic classifier, even if optimal
in the Bayes sense, is not guaranteed to provide a transformation that is
optimal for inference in any statistically meaningful way.

An alternative formulation of the limitations of classification for statistical
inference is based on the sufficiency conditions required for summary
statistics, according to the Fisher-Neyman factorisation criterion. A summary
statistic for a set of i.i.d. observations $D =
\{\boldsymbol{x}_0,...,\boldsymbol{x}_n\}$ is sufficient with respect to a
statistical model and a set of parameters $\boldsymbol{\theta}$ if and only the
generating probability distribution function of the data $p(\boldsymbol{x} |
\boldsymbol{\theta})$ can be factorised as follows:
\begin{equation}
p(\boldsymbol{x} | \boldsymbol{\theta}) =
                q(\boldsymbol{x})
                r(\boldsymbol{s}(\boldsymbol{x}) | \boldsymbol{\theta})
\label{eq:fisher_neyman_crit}
\end{equation}
where $q(\boldsymbol{x})$ is a non-negative function that does not depend on the
parameters and $r(\boldsymbol{x})$  is also a non-negative function for which
the dependence on the parameters $\boldsymbol{x}$ is a function of the summary
statistic $\boldsymbol{s}(\boldsymbol{x})$. Such a sufficient statistic contains
all the information in the observed sample useful for computing any estimate on the
model parameters, and no complementary statistic can add any additional
information about $\boldsymbol{\theta}$ contained in the set of observations.

A trivial sufficient summary statistic according to the previous definition is
the identity function $\boldsymbol{s}(\boldsymbol{x})=\boldsymbol{x}$, yet
typically we are only interested in summaries that reduce the original data
dimensionality. If $p(\boldsymbol{x} | \boldsymbol{\theta})$ is not known in
closed form, as is often the case in HEP analyses, the general
task of finding a sufficient summary statistic that reduces the dimensionality
cannot be tackled directly by analytic means. An exception to this can be easily
shown in the case of a mixture model where the mixture fraction $\mu$ is the only
parameter. By both dividing and multiplying by the mixture distribution function
from \eref{eq:mixture_structure} we easily obtain:
\begin{equation}
    p(\boldsymbol{x}| \mu, \boldsymbol{\theta} ) =
    p_b(\boldsymbol{x} | \boldsymbol{\theta})
    \left (
        1-\mu + \mu \frac{p_s(\boldsymbol{x} |
        \boldsymbol{\theta})}{p_b(\boldsymbol{x} | \boldsymbol{\theta})}
    \right )
\end{equation}
from which we can already prove that the density ratio $s_{s/
b}(\boldsymbol{x})= p_s(\boldsymbol{x} | \boldsymbol{\theta}) /
p_b(\boldsymbol{x} | \boldsymbol{\theta})$ (or alternatively its inverse) is a
sufficient summary statistic for the mixture coefficient parameter, according
to the Fisher-Neyman factorisation criterion from \eref{eq:fisher_neyman_crit}.
This quantity could be efficiently approximated by considering the problem of
probabilistic classification between signal and background as discussed in
\eref{eq:density_ratio_clf}. Because any bijective function of a sufficient
summary statistic is also a sufficient summary statistic, the conditional
probability from the conditional output of a balanced classifier
\begin{equation}
c(\boldsymbol{x})=s_{s/(s+b)} (\boldsymbol{x})= \frac{p_s(\boldsymbol{x} | \boldsymbol{\theta})}{
                p_s(\boldsymbol{x} | \boldsymbol{\theta}) +
                 p_b(\boldsymbol{x} | \boldsymbol{\theta})}
\end{equation}
can be used directly as a summary instead of $s_{s/ b}(\boldsymbol{x})$, with
the additional advantage that it is bounded between zero and one,
a fact that greatly simplifies visualisation and calibration.

From this perspective, the utility of signal versus background classification to
obtain an approximately sufficient summary statistic with respect to the mixture
model and mixture fraction $\mu$ is evident. However, if the statistical model
depends on additional nuisance parameters, even Bayes optimal probabilistic
classification does not provide any sufficient guarantees. Thus, even for
the best possible classifier that can be constructed,
useful information which can be used to constrain the parameters of interest might be
lost if a low-dimensional classification-based summary statistic is used in
place of the original data $\boldsymbol{x}$.

Above we have reviewed from a statistical perspective the limitations of signal versus background classification models when the goal is inference in the presence of nuisance parameters. In practice, classifiers can be trained for the most
probable likely value of the nuisance parameters and their effect can be
adequately accounted for during calibration, yet the resulting inference will be
degraded even if the classification is optimal. Alternative uses of
classification and regression models such as particle identification and momentum
or energy regression can be understood as approximations of a subset of relevant latent variables $\boldsymbol{z}$ of the generative model. This information could be then be used to complement the reconstruction output for each object and design better hand-crafted or classification-based summary statistics, so at the end the final goal is inference, and the previously mentioned shortcomings still apply.

\subsection{Generalising the likelihood ratio trick}
\label{sec:likelihood-ratio-general}

The first known attempts to study the relation between statistical inference in
HEP with nuisance parameters and probabilistic classifiers date
to Neal~\cite{Neal2008-om}. In his seminal paper, in addition of making explicit
the problem of not being able to compute the data generating likelihood in
closed form and clarifying the useful relation between likelihood ratios and
probabilistic classifiers discussed in the previous subsection, he also
acknowledges the limitations of this approach in the presence of nuisance
parameters and suggests a few possible candidate solutions.

To ameliorate the problems of losing useful information when reducing the
dimensionality of the data with summary statistics, a few variations over
classical signal versus background classifiers trained with the best
estimation of the nuisance parameters are proposed. The first proposal foresees the training of
a single robust classifier by combining simulated observations of signal and
background generated with different nuisance parameters, for example drawn from
their prior or from a reasonable distribution, to constrain the nuisance parameters $\pi(\boldsymbol{\theta})$.

The drawbacks of such marginal classifier are similar to the concerns about
models trained for the most likely values of the nuisance parameters: 
it might not be possible to accurately classify without knowing
$\boldsymbol{\theta}$, and even when that is possible the usefulness of the resulting score will be degraded when
calibrated statistical inference is carried out. To address these concerns, the
author suggests a generalisation based on training a single classifier
considering both the observations $\boldsymbol{x}$ and the nuisance parameters
$\boldsymbol{\theta}$ as input. The resulting model would be a nuisance-parameterized signal versus background classifier, thus an early precedent for some of the approaches discussed in \sref{sec:nuisance-parameterized-models}.
In order to use these parameterized classifiers on real data, for which the correct values of $\boldsymbol{\theta}$
are not known, Neal argues that an additional per-event regression model for
$\boldsymbol{\theta}$ could be trained on simulated observations.

The ideas developed by Neal~\cite{Neal2008-om} were not applied in practice
until they were generalised and extended by Cranmer \textit{et al.}~\cite{Cranmer2015-tq}. The authors of that cited work identify the same problem regarding the use of discriminative classifiers to approximate likelihood ratios with nuisance parameters and introduce a generic framework for inference using calibrated parameterized classifiers referred to as \textsc{Carl}. In their more general formulation, they propose using a doubly parameterized classifier to approximate the likelihood ratio for all possible pairs of relevant parameters
$\boldsymbol{\theta}_0$ and $\boldsymbol{\theta}_1$ of a generative model
$p(\boldsymbol{x} | \boldsymbol{\theta})$ as follows:
\begin{equation}
     \hat{r}(\boldsymbol{x}; \boldsymbol{\theta}_0, \boldsymbol{\theta}_1) \approx
    \frac{p(\boldsymbol{x}| \boldsymbol{\theta}_0)}{ p(\boldsymbol{x} | \boldsymbol{\theta}_1)}
\label{eq:likelihood_ratio_general_approx}
\end{equation}
where the classifier output $\hat{r}(\boldsymbol{x}; \boldsymbol{\theta}_0,
\boldsymbol{\theta}_1)$ has a specific dependence on the parameter vectors
$\boldsymbol{\theta}_0$ and $\boldsymbol{\theta}_1$ and the approximation
becomes an equality only for a Bayes optimal classifier for each combination. In
order to train such classifiers in a data-efficient manner, they suggest using
smooth models such as neural networks and a single learning stage based on a
large dataset where each observation correspond to an instantiation of the
parameters $\boldsymbol{\theta}_0$ and $\boldsymbol{\theta}_1$ drawn from a
reasonable prior distribution $\pi(\boldsymbol{\theta})$ and where
$\boldsymbol{x}$ is drawn from the generative model using those parameters.

Given a flexible enough model and enough training data, the procedure described above could be used to learn a good approximation of the quantity in
\eref{eq:likelihood_ratio_general_approx}. For problems where the underlying
structure is a mixture model, Cranmer \textit{et al.} also point out that is possible to obtain the quantity $\hat{r}(\boldsymbol{x}; \boldsymbol{\theta}_0, \boldsymbol{\theta}_1)$ based on the parameterized output
for each pairwise component classification problem which are simpler learning
tasks. Because, in practice, the approximation cannot be assumed to be exact, the
authors of \cite{Cranmer2015-tq} also propose to have a second stage where generative model samples are
used again to calibrate all the relevant values of the parameters as well as a
set of diagnostic procedures. They successfully apply this methodology to
a set of example problems and discuss it potential usefulness in the context of
HEP analysis.

\ifStandalone
It
\fi
is worth noting that the component of the
vector parameters in $\boldsymbol{\theta}$ in \textsc{Carl} could include both
nuisance parameters and parameters of interests in the same manner. The nuisance
parameters could also be incorporated in the calibration and profiled or marginalised
at the inference stage. Once we have a well-calibrated approximation of the likelihood
ratio, we can directly use it to construct arbitrary test statistics and confidence
intervals for statistical inference. Hence, with the caveats associated with a more involved
training procedure and parametric calibration procedure, this technique presents
the first principled and general solution for dealing with parameters when using
machine learning techniques in the context of HEP inference.

\subsection{Learning more efficiently from the simulator}
As mentioned earlier, one of the caveats of the general applicability of \textsc{Carl}
is that the training and probabilistic calibration\footnote{
Through this section by calibration we refer to the use of an independent set
of simulated data to transform the resulting estimator to ensure its
expected statistical properties. We refer to the calibration section
of~\cite{Brehmer:2018eca} for two different approaches to obtain this type
of transformations.} procedure may potentially require a large amount
of simulated data to approximate accurately the likelihood ratio
$r(\boldsymbol{x}; \boldsymbol{\theta}_0, \boldsymbol{\theta}_1)$ for all
relevant $\boldsymbol{\theta}_0$ and $\boldsymbol{\theta}_1$ when the dimension
of $\boldsymbol{\theta}$. This practical limitation motivated Brehmer and the
original authors of \textsc{Carl} to develop a family of
methods~\cite{Brehmer:2018hga,Brehmer:2018kdj,Brehmer:2018eca} to estimate the
likelihood ratio and other useful quantities for inference in a more data-efficient manner,
by augmenting training data with information from the
simulator. The source of the information from the simulations comes from the
properties and structure of the data generating process:
\begin{equation}
    p ( \boldsymbol{x}|\boldsymbol{\theta} ) =
                \int p(\boldsymbol{x}, \boldsymbol{z} | \boldsymbol{\theta})d\boldsymbol{z}
\end{equation}
which are characterised by the the joint distribution function
$p(\boldsymbol{x}, \boldsymbol{z} | \boldsymbol{\theta})d\boldsymbol{z}$ where
$\boldsymbol{z}$ are all the latent variables of each observation. In
high-energy physics event generation, the joint probability distribution can be
factorised in a series of conditional distributions matching the various simulation
steps and their dependencies:
\begin{equation}
    p(\boldsymbol{x}, \boldsymbol{z} | \boldsymbol{\theta}) =
                p ( \boldsymbol{x} | \boldsymbol{z}_\textrm{d})
                p ( \boldsymbol{z}_\textrm{d} | \boldsymbol{z}_\textrm{s})
                p ( \boldsymbol{z}_\textrm{s} | \boldsymbol{z}_\textrm{p})
                \sum^{K-1}_{j=0} p ( z_i  = j |\boldsymbol{\theta}_\textrm{th})
                p ( \boldsymbol{z}_\textrm{p}|\boldsymbol{\theta}_\textrm{th}, z_i  = j)
\end{equation}
where $p( z_i = j|\boldsymbol{\theta})$ is the probability of a given type of
process $j$ occurring, $p ( \boldsymbol{z}_\textrm{p}|\boldsymbol{\theta}, z_i =
j)$ is the conditional probability density of a given set of parton-level
four-momenta particles for a given process, $p( \boldsymbol{z}_\textrm{s} |
\boldsymbol{z}_\textrm{p})$ is the conditional density of a given parton-shower
outcome, $p ( \boldsymbol{z}_\textrm{d} | \boldsymbol{z}_\textrm{s})$ is the
conditional density of a set of detector interactions and readout noise and $p
( \boldsymbol{x} | \boldsymbol{z}_\textrm{d})$ is the conditional density of a
given detector readout. Note that all the factors could depend on additional
nuisance parameters; here only the theoretical parameters $\boldsymbol{\theta}_\textrm{th}$ are made
explicit for notational simplicity because they are normally the parameters of
interest. Also note that the last factor gives rise to the mixture structure mentioned in the last subsection. While $p ( \boldsymbol{x}|\boldsymbol{\theta})$ and ratios of that quantity are typically intractable, the authors suitably remark that the joint likelihood ratio
\begin{equation}
    r(\boldsymbol{x}, \boldsymbol{z} | \boldsymbol{\theta}_0, \boldsymbol{\theta}_1) =
    \frac{p(\boldsymbol{x}, \boldsymbol{z} | \boldsymbol{\theta}_0)}{
    p(\boldsymbol{x}, \boldsymbol{z} | \boldsymbol{\theta}_1)}
\end{equation}
and the joint score
\begin{equation}
    t(\boldsymbol{x}, \boldsymbol{z} | \boldsymbol{\theta}_0) = \left.
    \nabla_{\boldsymbol{\theta}} \log
    p(\boldsymbol{x}, \boldsymbol{z} | \boldsymbol{\theta}) \right|_{\boldsymbol{\theta}_0}
\end{equation}
can often be obtained exactly for a given simulated observation due to its factorised structure. They propose two regression losses $L_r$
and $L_b$ for each of the previous quantities, which may be used to obtain an
approximation of the likelihood ratio $r(\boldsymbol{x}| \boldsymbol{\theta}_0,
\boldsymbol{\theta}_1)$ and the score $t(\boldsymbol{x}| \boldsymbol{\theta}_0)$
by empirical risk minimisation with various machine learning models such as
neural networks. Based on these loss functions, Brehmer \textit{et al.} develop a family of new methods as well as extensions of \textsc{Carl} to more efficiently
approximate the parameterized likelihood ratio $r(\boldsymbol{x}|
\boldsymbol{\theta}_0, \boldsymbol{\theta}_1)$ and demonstrate their
effectiveness in a few example problems. Another practical innovation developed
by the authors, applicable to all the new parameterized likelihood ratio
estimators and also to \textsc{Carl}, is that the parameters of the reference
hypothesis in $\boldsymbol{\theta}_1$ in
\eref{eq:likelihood_ratio_general_approx} can be kept fixed at an arbitrary
value, thus simplifying the learning task significantly. Building upon this
work, Stoye and the previous authors~\cite{Stoye:2018ovl} also developed two new
methods that can incorporate the joint likelihood ratio and the joint score
to a loss function based on the cross entropy, which reduces the variance during
the learning tasks further improving sample efficiency for obtaining accurate
likelihood ratio approximations.

In addition to the efficient techniques for parameterized likelihood ratio estimation discussed above,
Brehmer \textit{et al.}~\cite{Brehmer:2018hga,Brehmer:2018kdj,Brehmer:2018eca} also
propose a new class of methods referred to as \textsc{Sally} using the regressed
score approximation $\hat{t}(\boldsymbol{x}| \boldsymbol{\theta}_\textrm{ref})$
at a single reference parameter point $\boldsymbol{\theta}_\textrm{ref}$ to
construct a summary statistic. The score $t(\boldsymbol{x}|
\boldsymbol{\theta}_\textrm{ref})$, whose dimensionality is the same as that of
the parameter vector $\boldsymbol{\theta}$, is a sufficient statistic in the
neighborhood of $\boldsymbol{\theta}_\textrm{ref}$, so it is a very useful
transformation. Because the dimensionality might still be high in cases with a
large number of parameters, they propose to use a one-dimensional
projection (i.e. \textsc{Sallino}) in the direction of parameter variation as
alternative lower-dimensional statistic. In the same work, the authors also experiment with augmenting conditional neural density estimators, such as density networks or normalizing flows, with a
joint score regression loss function. A calibrated
estimation of the likelihood
$\hat{p}( \boldsymbol{x}|\boldsymbol{\theta})$ can be used as basis for any
statistical inference task but its accurate approximation is challenging with a
finite data sample, yet many recent advances coming from the field of machine
learning in density estimation could eventually make this approach viable.

Similarly to \textsc{Carl}, all these improved techniques for the estimation of
likelihood ratios, likelihood scores or the conditional likelihood itself make
no distinction between the statistical parameters in the model. Hence,
nuisance parameters can be incorporated in the vector of parameters
$\boldsymbol{\theta}$, accounted for like any another parameter in the
calibration, and profiled or marginalised at the inference stage. The challenge for their
direct application in HEP, particularly for the methods that use
augmented data from the simulator, is to approximate or model the effect of all
relevant nuisance parameters in the joint likelihood ratio and score. In a
recent publication, Brehmer \textit{et al.}~\cite{Brehmer:2019xox} presented a software
library to simplify the application of these techniques to LHC measurements and
included the effect of nuisance parameters from scale and parton distribution
function choices by varying the weights associated to each simulated
observations.



\subsection{Inference-aware summary statistics}

The previous techniques, with the exception of \textsc{Sally} (and
\textsc{Sallino} for a fixed projection) are based directly on calibrated
likelihood ratios or likelihood approximations so they are at their core a
different form of inference from what is typically done in HEP: they are designed to tackle the inference problem directly, rather than to
construct summary statistics. Such a strong paradigm change can be very
advantageous but also poses some challenges for its adoption. In recent years,
another complementary family of inference-aware techniques has been proposed,
whose objective is the construction of machine-learning based summary statistics
that are better aligned with the statistical inference goal of HEP analysis, including nuisance parameters. Once constructed, these summary
statistics can be used in place of simplified physical summaries or signal
background classification outputs.

A generic technique in this category, which has direct applicability to
HEP analyses, is \textsc{inferno}~\cite{De_Castro2019-pn}. In
that work, authors demonstrate how non-linear summary statistics can be
constructed by minimising inference-motivated losses via stochastic gradient
descent specific for the analysis goal. For example, for an analysis focusing on
the measuring of a physical quantity such as a cross section, the proposed approach
can be used to minimise directly, as a loss, an approximation of the excepted
uncertainty on the parameter of interest, fully accounting for the effect of
relevant nuisance parameters.

\begin{figure}
\centerline{\includegraphics[width=\linewidth]{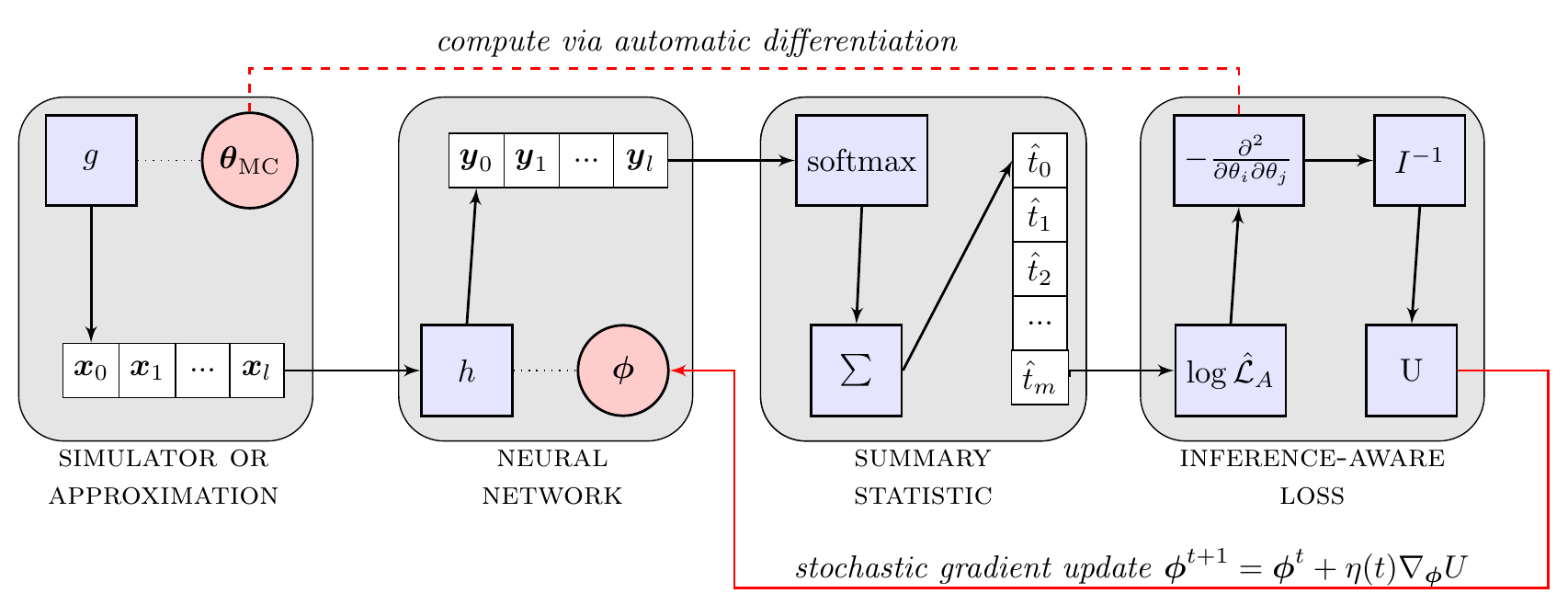}}
\caption{Learning inference-aware summary statistics. Figure by the authors of
\textsc{inferno}\cite{De_Castro2019-pn} and licensed under CC BY 4.0.}
\label{fig:figure1-inferno-paper}
\end{figure}

In \textsc{inferno} and other similar approaches discussed later,
the parameters of a neural network are optimised by stochastic gradient descent
within an automatic differentiation framework, where the considered loss
function accounts for the details of the statistical model as well as the
expected effect of nuisance parameters. A graphical depiction of this technique
is included in \fref{fig:figure1-inferno-paper}. The left-most block refers to
sampling a differentiable simulator or approximating the effect of the
parameters $\boldsymbol{\theta}$ over existing simulated observations, including
relevant nuisance parameters. These observations go through a neural network
that depends on a set of parameters $\boldsymbol{\phi}$ (second block from the
left) and then a histogram-like summary statistics are constructed from the
neural network output (third block from the left). Still within the automatic
differentiation framework, a synthetic likelihood (e.g. product of Poisson
counts for a histogram-like summary statistic) is constructed. A final
inference-aware loss, for example an approximation of the expected uncertainty
for the parameters of interest accounting for nuisance parameters, can then be
constructed based on the inverse Hessian matrix and used to optimise the neural
network parameters.

Note that the approximations used to make a differentiable
loss (e.g. continuous relaxation of a histogram) do not affect the
rigour of the resulting statistical inference.
Once the summary statistic transformation has been learned with the
procedure described above, it can be used, e.g. using an argmax operator instead of a softmax to compute the summary statistic if the approximation of
\fref{fig:figure1-inferno-paper} is used, to carry out statistical inference
with the with usual procedures and tools as would be done for any other histogram-based
summary statistic. The main challenge of using this
approach in HEP analyses is that the effect of nuisance parameters has
to be included in the auto-differentiation framework, for example by
transforming the input features (e.g. momenta and energy calibration
uncertainties), by interpolating simulated observation weights (e.g. theoretical
and parton distribution function uncertainties) or by considering the
interpolation between histogram counts as a last resource. If those challenges
can be overcome (even just for part of the nuisance parameters), this method
provides an alternative to perform dimensionality
reduction using directly an approximation of the inference objective of a given
analysis, in contrast with a
transformation based on probabilistic classification or a physics-motivated
feature. The authors demonstrate the effectiveness of this technique in a
multi-dimensional synthetic example with up to 3 nuisance parameters, where the
inference-aware summary statistics outperform even optimal classification-based
summaries.

A technique with a similar reach, but that was developed instead for tackling
likelihood-free inference problems in astrophysical observations, was presented
by Charnock \textit{et al.}~\cite{Charnock2018-bl}. In their work, the authors propose
information-maximising neural networks (\textsc{imnn}), a machine learning
technique to find non-linear functionals of the data that maximise the
the Fisher information. The Fisher information during  training
is computed from the Fisher matrix determinant, that it is itself
calculated from the derivatives of the outputs of the network with respect to
the parameters of inference at fiducial values by numerical differentiation
or directly from the adjoint gradient of a large number of simulations.
The authors additionally
propose the inclusion of the determinant of the covariance matrix of
the neural network outputs in the loss to control the magnitude of the summaries. While
they do not consider the problem of nuisance parameters specifically,
their approach will by design find transformations that are minimally affected
by nuisance parameters while being maximally sensitive to the parameters.
On a related note,
Alsing \textit{et al.}~\cite{Alsing2019-mp} develop a useful transformation that can be
applied to implicitly marginalise the summary statistics resulting from
\textsc{imnn} or score $t(\boldsymbol{x} | \boldsymbol{\theta})=
\nabla_{\boldsymbol{\theta}}\log p(\boldsymbol{x} | \boldsymbol{\theta})$
approximations (e.g. \textsc{Sally} from the previous subsection).

More recently, there has also been some recent work building upon the ideas
behind \textsc{inferno} that attempt to simplify its application to high-energy
physics analysis or extend its functionality. For example, Wunch et
al.~\cite{Wunsch2020-yw} suggest using a differentiable transformation of a
neural network with a single node to construct a Poisson count likelihood instead
of a softmax as the basis for the inference-aware loss. Similarly to what was
observed for \textsc{inferno}, the authors demonstrate the usefulness of an
inference-aware construction in a synthetic example, and also using an extension
of the Higgs ML benchmark including nuisance parameters. Following a different
path, the authors of \textsc{neos}~\cite{lukas_heinrich_2020_3697981} use a
technique referred to as fixed-point differentiation to compute gradients of the
profile likelihood, thus avoiding the Hessian inverse approximation, and to
directly minimise the expected upper limits $\textrm{CL}_s$. Both Wunch \textit{et al.}
and the authors of \textsc{neos} at the time restrict the modelling of the effect of
nuisance parameters to histogram interpolation.

In addition to the mentioned approaches, it is worth noting other alternatives
with a more limited range of applicability but that could be useful for certain
use cases. Elwood \textit{et al.}~\cite{Elwood2018-jn} propose using the expected
significance approximation formula for a single bin count experiment, optionally
including the effect of a single source of systematic uncertainty directly as a
loss of a neural network. For a different type of model, Xia~\cite{Xia2018-xf}
develop a variation of boosted decision tree training referred to as
\textsc{qbdt} which targets directly the statistical significance, and which can also include the effect of nuisance parameters in its approximation. In both cases, authors demonstrate with practical examples that the significance optimising algorithms outperform their classification counterparts.

\section{Outlook}

The reduction of the effect of systematic uncertainties in parameter estimation
is a crucial problem in particle physics. In the past, the problem was attacked by striving for redundancy of the measurement apparata, robustness of the detection techniques, and the use of analysis methods aiming for inter-calibration, cross-validation, and leveraging as much as possible control datasets and measurements. In the machine learning era, automated methods have become available that may significantly further reduce the impact that imprecise knowledge of latent features of the data have on physics measurements. While already a significant arsenal of techniques has been amassed, no catch-all procedure has emerged yet, so insight is still required to discern the salient features of the problem to be solved and the appropriate method to deploy. The most promising avenues for a general procedure of handling nuisance parameters are those described in \sref{sec:inference-aware-approaches}, where the optimization objectives are more directly linked to the inference goal.

\section*{Acknowledgements}

The authors would like to thank Johann Brehmer for some discussions about the techniques
included in \textsc{MadMiner} and how they can deal with nuisance parameters. The authors
would also like to thank all the anonymous reviewers of this review for their
feedback and comments.
\clearpage
\clearpage

\bibliographystyle{tepml}
\bibliography{ws-rv-sample}


\end{document}